\RequirePackage{fix-cm}
\documentclass[twocolumn]{svjour3}       
\smartqed  
\usepackage{xcolor}

\usepackage{graphicx}
\usepackage{amsmath}
\usepackage{pdflscape}
\usepackage{tabularx}
\usepackage{lipsum}
\usepackage{multicol}
\usepackage{multirow}
\usepackage{array}
\usepackage{makecell}
\usepackage{caption}
\usepackage{CJK}
\PassOptionsToPackage{hyphens}{url}\usepackage{hyperref}
\hypersetup{
    colorlinks=true,
    linkcolor=blue,
    filecolor=blue,      
    urlcolor=blue,
    citecolor=blue,
}
\usepackage{natbib}
\setcitestyle{authoryear,open={(},close={)}}

\usepackage{verbatim}
\usepackage{float}
\usepackage{amssymb}

\begin{document}

\title{Scene Text Detection and Recognition:
}
\subtitle{The Deep Learning Era}


\author{Shangbang Long \and
        Xin He \and 
        Cong Yao
}


\institute{Shangbang Long \at
              Machine Learning Department, School of Computer Science \\ 
              Carnegie Mellon University \\
              \email{shangbal@cs.cmu.edu}           
           \and
           Xin He \at
           ByteDance Ltd. \\
          \email{hexin7257@gmail.com}
            \and
           Cong Yao \at
           MEGVII Inc. (Face++) \\ 
           \email{yaocong2010@gmail.com}
}

\date{Received: date / Accepted: date}

\maketitle

\begin{abstract}
With the rise and development of deep learning, computer vision has been tremendously transformed and reshaped. As an important research area in computer vision, scene text detection and recognition has been inevitably 
influenced by this wave of revolution, consequentially entering the era of deep learning. In recent years, the community has witnessed substantial advancements in mindset, methodology and performance. This survey is aimed at summarizing and analyzing the major changes and significant progresses of scene text detection and recognition in the deep learning era. Through this article, we devote to: (1) introduce new insights and ideas; (2) highlight recent techniques and benchmarks; (3) look ahead into future trends. Specifically, we will emphasize the dramatic differences brought by deep learning and remaining grand challenges. We expect that this review paper would serve as a reference book for researchers in this field. Related resources are also collected in our Github repository \footnote{\url{https://github.com/Jyouhou/SceneTextPapers}}.
\end{abstract}

\section{Introduction}\label{sec:introduction}

Undoubtedly, text is among the most brilliant and influential creations of humankind. As the written form of human languages, text makes it feasible to reliably and effectively spread or acquire information across time and space. In this sense, text constitutes the cornerstone of human civilization. 

\begin{figure}
\centering
\includegraphics[width=1.0\columnwidth]{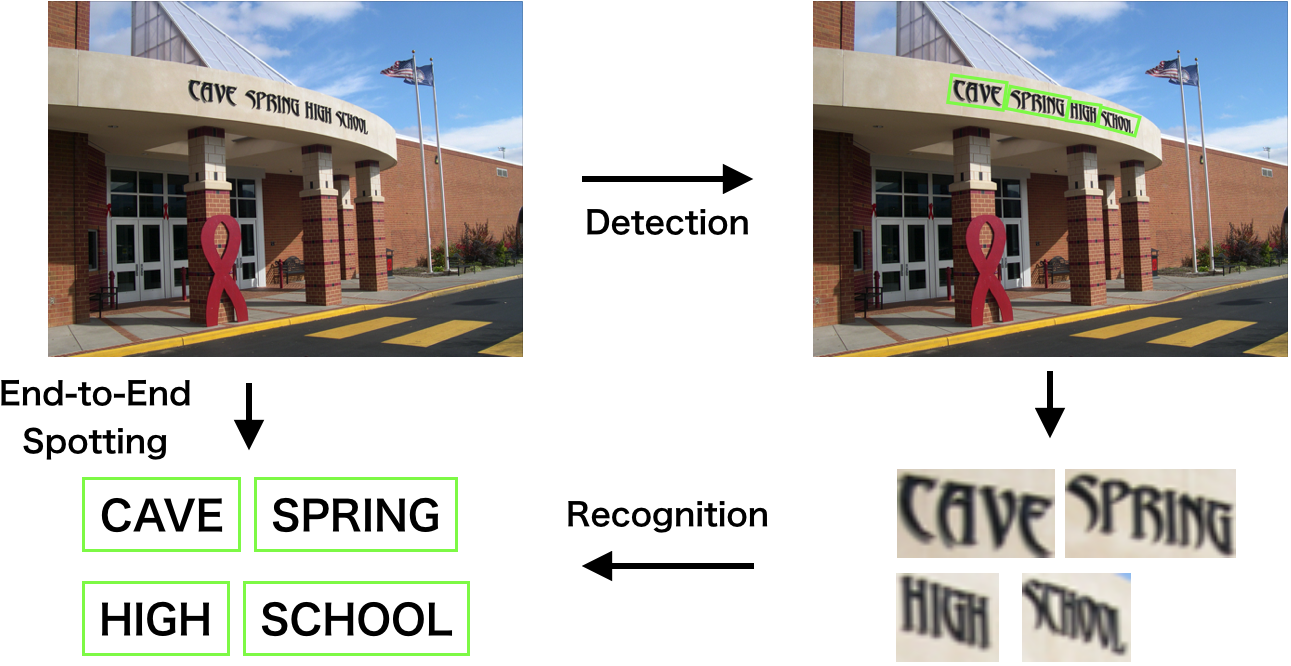}
\caption{Schematic diagram of scene text detection and recognition. The image sample is from Total-Text~\citep{kheng2017total}.}  
\label{fig:concept}
\end{figure}

On the one hand, as a vital tool for communication and collaboration, text has been playing a more important role than ever in modern society; on the other hand, the rich and precise high-level semantics embodied in text could be beneficial for understanding the world around us. For example, text information can be used in a wide range of real-world applications, such as \textit{image  search}~\citep{tsai2011mobile,schroth2011exploiting}, \textit{instant translation}~\citep{dvorin2009method,parkinson2016instant}, \textit{robots navigation}~\citep{desouza2002vision,liu2005edge,liu2005simple,schulz2015robot}, and \textit{industrial automation}~\citep{ham1995recognition,he2005new,chowdhury2013extracting}. Therefore, automatic text reading from natural environments, as illustrated in Fig.~\ref{fig:concept}, a.k.a. scene text detection and recognition~\citep{zhu2016scene} or PhotoOCR~\citep{bissacco2013photoocr}, has become an increasing popular and important research topic in computer vision. 

However, despite years of research, a series of grand challenges may still be encountered when detecting and recognizing text in the wild. The difficulties mainly stem from three aspects:

\noindent $\bullet$ \textbf{Diversity and Variability of Text in Natural Scenes} Distinctive from scripts in documents, text in natural scene exhibit much higher diversity and variability. For example, instances of scene text can be in different languages, colors, fonts, sizes, orientations, and shapes. Moreover, the aspect ratios and layouts of scene text may vary significantly. All these variations pose challenges for detection and recognition algorithms designed for text in natural scenes.

\noindent $\bullet$ \textbf{Complexity and Interference of Backgrounds} The backgrounds of natural scenes are virtually unpredictable. There might be patterns extremely similar to text (e.g., tree leaves, traffic signs, bricks, windows, and stockades), or occlusions caused by foreign objects, which may potentially lead to confusion and mistakes.

\noindent $\bullet$ \textbf{Imperfect Imaging Conditions} In uncontrolled circumstances, the quality of text images and videos could not be guaranteed. That is, in poor imaging conditions, text instances may be of low resolution and severe distortion due to inappropriate shooting distance or angle, or blurred because of out of focus or shaking, or noised on account of low light level, or corrupted by highlights or shadows.

\begin{figure}
\centering
\includegraphics[width=1.0\columnwidth]{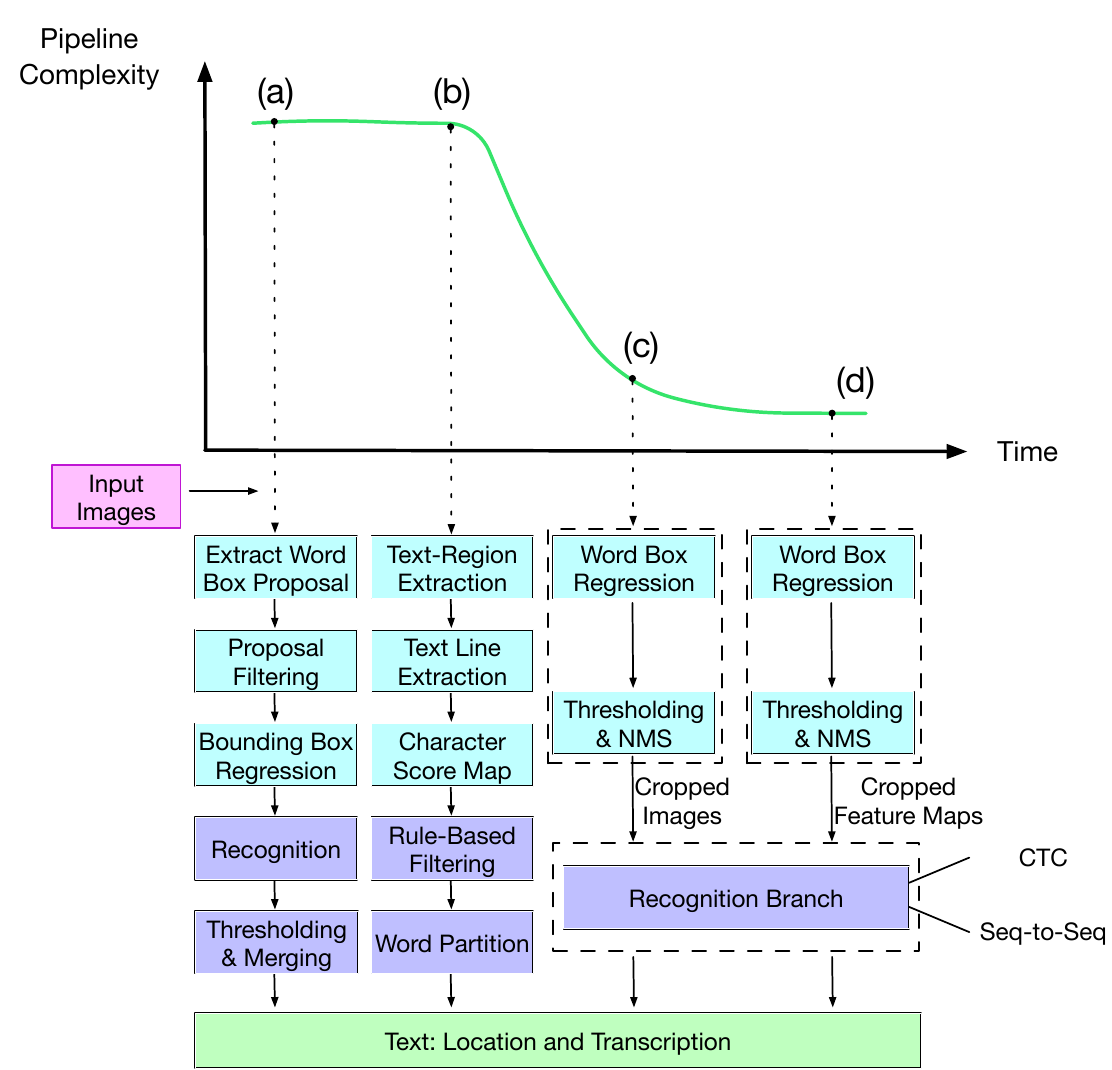}
\caption{Illustrations of representative scene text detection and recognition system pipelines.  {(a)}~\cite{jaderberg2016reading} and {(b)}~\cite{yao2016scene} are representative multi-step methods. {(c)} and {(d)} are simplified pipeline. 
In {(c)}, detectors and recognizers are separate. 
In {(d)}, the detectors pass cropped feature maps to recognizers, which allows end-to-end training.}  
\label{fig:pipeline}
\end{figure}

These difficulties run through the years before deep learning showed its potential in computer vision as well as in other fields. As deep learning came to prominence after AlexNet~\citep{krizhevsky2012imagenet} won the ILSVRC2012~\citep{russakovsky2015imagenet} contest, researchers turn to deep neural networks for automatic feature learning and start with more in-depth studies. The community are now working on ever more challenging targets. The progress made in recent years can be summarized as follows:

\noindent $\bullet$ \textbf{Incorporation of Deep Learning} Nearly all recent methods are built upon deep learning models. Most importantly, deep learning frees researchers from the exhausting work of repeatedly designing and testing hand-crafted features, which gives rise to a blossom of works that push the envelope further. To be specific, the use of deep learning substantially simplifies the overall pipeline, as illustrated in Fig. \ref{fig:pipeline}. Besides, these algorithms provide significant improvements over previous ones on standard benchmarks. Gradient-based training routines also facilitate to end-to-end trainable methods.

\noindent $\bullet$ \textbf{Challenge-Oriented Algorithms and Datasets} Researchers are now turning to more specific aspects and challenges. 
Against difficulties in real-world scenarios, newly published datasets are collected with unique and representative characteristics. 
For example, there are datasets featuring long text~\citep{tu2012detecting}, blurred text~\citep{karatzas2015icdar}, and curved text~\citep{kheng2017total} respectively. 
Driven by these datasets, almost all algorithms published in recent years are designed to tackle specific challenges. 
For instance, some are proposed to detect oriented text, while others aim at blurred and unfocused scene images. 
These ideas are also combined to make more general-purpose methods.

\noindent $\bullet$ \textbf{Advances in Auxiliary Technologies} Apart from new datasets and models devoted to the main task, auxiliary technologies that do not solve the task directly also find their places in this field, such as synthetic data and bootstrapping.

In this survey, we present an overview of the recent development in deep-learning-based text detection and recognition from still scene images. We review methods from different perspectives and list the up-to-date datasets. We also analyze the status quo and future research trends.

There have been already several excellent review papers~\citep{uchida2014text,ye2015text,yin2016text,zhu2016scene}, which also organize and analyze works related to text detection and recognition. However, these papers are published before deep learning came to prominence in this field. Therefore, they mainly focus on more traditional and feature-based methods. We refer readers to these paper as well for a more comprehensive view and knowledge of the history of this field. This article will mainly concentrate on text information extraction from still images, rather than videos. For scene text detection and recognition in videos, please also refer to \cite{jung2004text} and \cite{yin2016text}.

The remaining parts of this paper are arranged as follows: In Section \ref{sec-2}, we briefly review the methods before the deep learning era. 
In Section \ref{sec-3}, we list and summarize algorithms based on deep learning in a hierarchical order. 
Note that we do not introduce these techniques in a paper-by-paper order, but instead based on a taxonomy of their methodologies. 
Some papers may appear in several sections if they have contributions to multiple aspects. 
In Section \ref{sec-4}, we take a look at the datasets and evaluation protocols. 
Finally, in Section \ref{sec-5} and Section \ref{sec-6}, we present potential applications and our own opinions on the current status and future trends.

\section{Methods before the Deep Learning Era}\label{sec-2}
In this section, we take a glance retrospectively at algorithms before the deep learning era. More detailed and comprehensive coverage of these works can be found in~\citep{uchida2014text,ye2015text,yin2016text,zhu2016scene}. For text detection and recognition, the attention has been the design of features. 

\begin{figure}
\centering
\includegraphics[width=0.8\columnwidth]{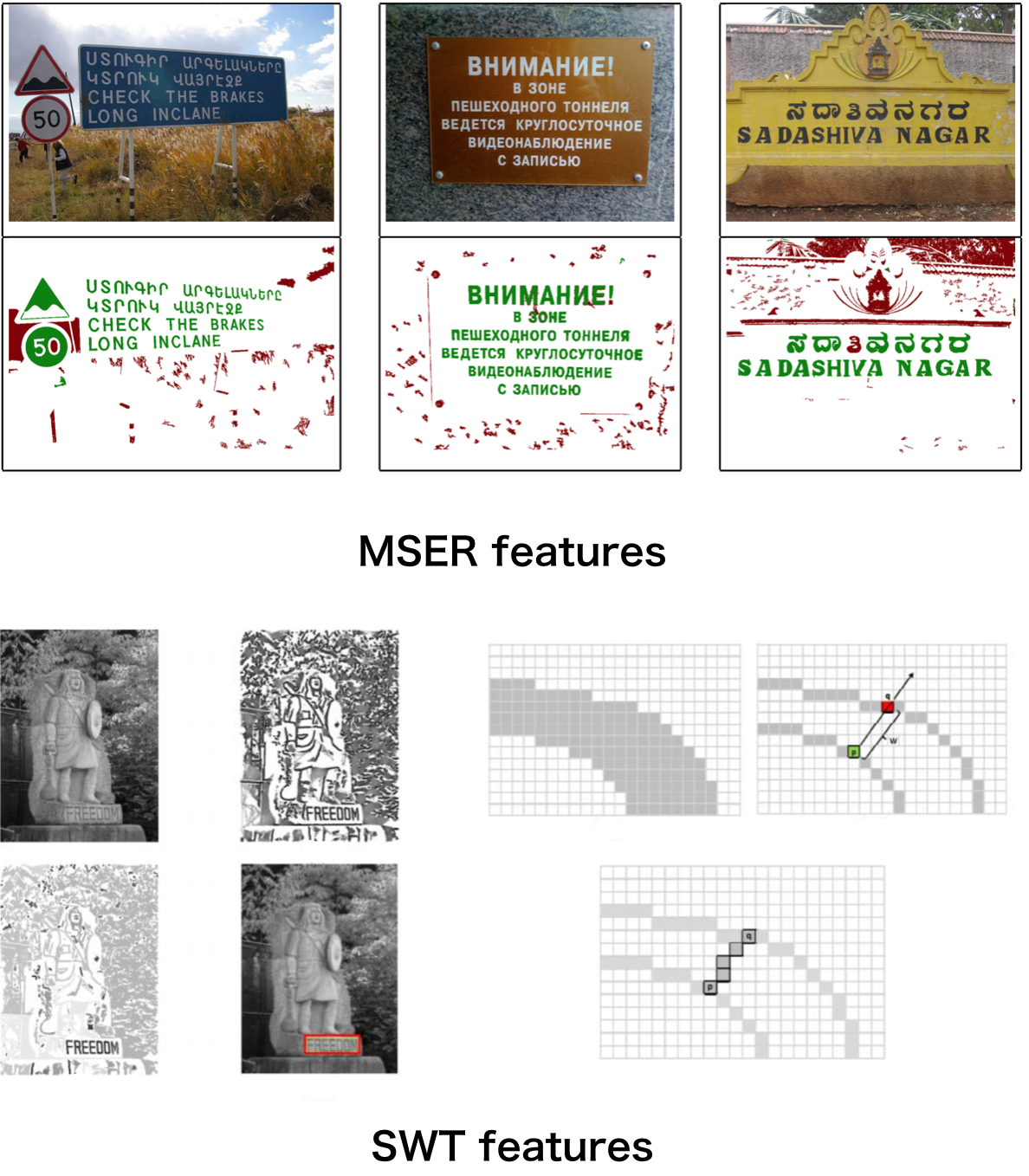}
\caption{Illustration of traditional methods with hand-crafted features: (1) {Maximally Stable Extremal Regions} (MSER)~\citep{neumann2010method}, assuming chromatic consistency within each character; (2) {Stroke Width Transform} (SWT)~\citep{epshtein2010detecting}, assuming consistent stroke width within each character. 
}  
\label{fig:traditional_methods}
\end{figure}

In this period of time, most text detection methods either adopt {Connected Components Analysis} (CCA) ~\citep{huang2013text,neumann2010method,epshtein2010detecting,tu2012detecting,yin2014robust,yi2011text,jain1998automatic} or {Sliding Window} (SW) based classification~\citep{lee2011adaboost,wang2011end,coates2011text,wang2012end}.  CCA based methods first extract candidate components through a variety of ways (e.g., color clustering or extreme region extraction), and then filter out non-text components using manually designed rules or classifiers automatically trained on hand-crafted features (see Fig.\ref{fig:traditional_methods}). In sliding window classification methods,  windows of varying sizes slide over the input image, where each window is classified as text segments/regions or not. Those classified as positive are further grouped into text regions with morphological operations~\citep{lee2011adaboost}, {Conditional Random Field} (CRF)~\citep{wang2011end} and other alternative graph based methods~\citep{coates2011text,wang2012end}.

For text recognition, one branch adopted the feature-based methods. \cite{shi2013scene} and \cite{yao2014strokelets}  propose {character segments} based recognition algorithms. 
\cite{rodriguez2013label,rodriguez2015label,gordo2015supervised,almazan2014word} utilize {label embedding} to directly perform matching between strings and images. {Strokes}~\citep{busta2015fastext} and {character key-points} ~\citep{quy2013recognizing} are also detected as features for classification. Another decomposes the recognition process into a series of sub-problems. Various methods have been proposed to tackle these {sub-problems}, which includes text binarization~\citep{zhiwei2010edge,mishra2011mrf,wakahara2011binarization,lee2013integrating}, text line segmentation~\citep{ye2003robust}, character segmentation~\citep{nomura2005novel,shivakumara2011new,roy2009multi}, single character recognition~\citep{chen2004automatic,sheshadri2012exemplar} and word correction~\citep{zhang2003bayesian,wachenfeld2006recognition,mishra2012scene,karatzas2004text,weinman2007fast}. 

There have been efforts devoted to integrated (i.e. end-to-end as we call it today) systems as well~\citep{wang2011end,neumann2013combining}. In Wang \emph{et al.}~\citep{wang2011end}, characters are considered as a special case in object detection and detected by a nearest-neighbor classifier trained on HOG features~\citep{dalal2005histograms} and then grouped into words through a Pictorial Structure (PS) based model~\citep{felzenszwalb2005pictorial}. Neumann and Matas~\citep{neumann2013combining} proposed a decision delay approach by keeping multiple segmentations of each character until the last stage when the context of each character is known. They detect character segmentation using extremal regions and decode recognition results through a dynamic programming algorithm.

In summary, text detection and recognition methods before the deep learning era mainly extract low-level or mid-level handcrafted image features, which entails demanding and repetitive pre-processing and post-processing steps. Constrained by the limited representation ability of handcrafted features and the complexity of pipelines, those methods can hardly handle intricate circumstances, e.g. blurred images in the ICDAR 2015 dataset~\citep{karatzas2015icdar}.

\section{Methodology in the Deep Learning Era}\label{sec-3}

As implied by the title of this section, we would like to address recent advances as changes in \textit{methodology} instead of merely new \textit{methods}. Our conclusion is grounded in the observations as explained in the following paragraph.

Methods in the recent years are characterized by the following two distinctions: (1) Most methods utilize deep-learning based models; (2) Most researchers are approaching the problem from a diversity of perspectives, trying to solve different challenges. 
Methods driven by deep learning enjoy the advantage that automatic feature learning can save us from designing and testing a large amount of potential hand-crafted features. At the same time, researchers from different viewpoints are enriching and promoting the community into more in-depth work, aiming at different targets, e.g. faster and simpler pipeline~\citep{Zhou_2017_CVPR}, text of varying aspect ratios~\citep{Shi_2017_CVPR}, and synthetic data~\citep{gupta2016synthetic}. As we can also see further in this section, the incorporation of deep learning has totally changed the way researchers approach the task and has enlarged the scope of research by far. This is the most significant change compared to the former epoch.

In this section, we would classify existing methods into a hierarchical taxonomy, and introduce them in a top-down style. First, we divide them into four kinds of systems: (1) text detection that detects and localizes text in natural images; (2) recognition system that transcribes and converts the content of the detected text regions into linguistic symbols; (3) end-to-end system that performs both text detection and recognition in one unified pipeline; (4) auxiliary methods that aim to support the main task of text detection and recognition, e.g. synthetic data generation. Under each category, we review recent methods from different perspectives.

\subsection{Detection}

We acknowledge that scene text detection can be taxonomically subsumed under general object detection, which is dichotomized as one-staged methods and two-staged ones. 
Indeed, many scene text detection algorithms are majorly inspired by and follow the designs of general object detectors. 
Therefore we also encourage readers to refer to recent surveys on object detection methods~\citep{han2018advanced,liu2018deep}.
However, the detection of scene text has a different set of characteristics and challenges that require unique methodologies and solutions. 
Thus, many methods rely on special representation for scene text to solve these non-trivial problems. 

The evolution of scene text detection algorithms, therefore, undergoes three main stages: 
(1) In the first stage, learning-based methods are equipped with multi-step pipelines, but these methods are still slow and complicated. 
(2) Then, the idea and methods of general object detection are successfully implanted into this task. 
(3) In the third stage, researchers design special representations based on sub-text components to solve the challenges of long text and irregular text.

\subsubsection{Early Attempts to Utilize Deep Learning}
Early deep-learning-based methods
~\citep{huang2014robust,tian2015text,yao2016scene,zhang2016multi,he2017multi} approach the task of text detection into a multi-step process. 
They use convolutional neural networks (CNNs) to predict local segments and then apply heuristic post-processing steps to merge segments into detection lines.

In an early attempt~\citep{huang2014robust}, CNNs are only used to classify local image patches into text and non-text classes. 
They propose to mine such image patches using MSER features. 
Positive patches are then merged into text lines. 

Later, CNNs are applied to the whole images in a fully convolutional approach. 
TextFlow~\citep{tian2015text} uses CNNs to detect character and views the character grouping task as a min-cost flow problem~\citep{goldberg1997efficient}.

In~\citep{yao2016scene}, 
a convolutional neural network is used to predict whether each pixel in the input image (1) belongs to characters, (2) is inside the text region, and (3) the text orientation around the pixel. 
Connected positive responses are considered as detected characters or text regions. For characters belonging to the same text region, Delaunay triangulation~\citep{kang2014orientation} is applied, after which a graph partition algorithm groups characters into text lines based on the predicted orientation attribute. 

Similarly, ~\cite{zhang2016multi} first predicts a segmentation map indicating text line regions. For each text line region, MSER~\citep{neumann2012real} is applied to extract character candidates. Character candidates reveal information on the scale and orientation of the underlying text line. Finally, minimum bounding boxes are extracted as the final text line candidates.

~\cite{he2017multi} propose a detection process that also consists of several steps. First, text blocks are extracted. 
Then the model crops and only focuses on the extracted text blocks to extract text center line (TCL), which is defined as a shrunk version of the original text line. 
Each text line represents the existence of one text instance. 
The extracted TCL map is then split into several TCLs. 
Each split TCL is then concatenated to the original image. 
A semantic segmentation model then classifies each pixel into ones that belong to the same text instance as the given TCL, and ones that do not. 

Overall, in this stage, scene text detection algorithms still have long and slow pipelines, though they have replaced some hand-crafted features with learning-based ones. 
The design methodology is bottom-up and based on key components, such as single characters and text center lines.

\subsubsection{Methods Inspired by Object Detection}
\begin{figure}
\centering
\includegraphics[width=1.0\columnwidth]{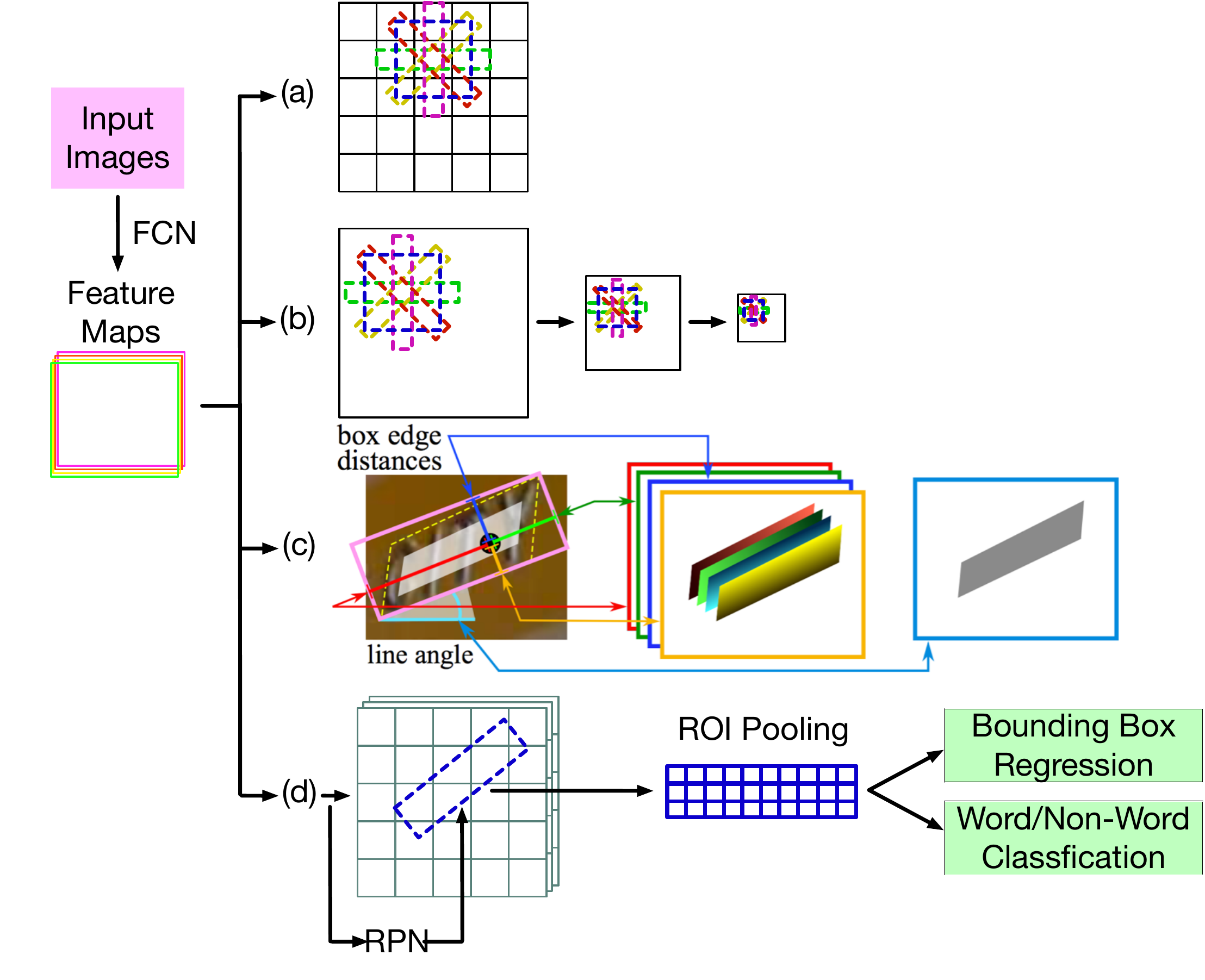}
\caption{High-level illustration of methods inspired by general object detection: (a) Similar to YOLO~\citep{redmon2016you}, regressing offsets based on default bounding boxes at each anchor position. 
(b) Variants of SSD~\citep{liu2016ssd}, predicting at feature maps of different scales. 
(c) Predicting at each anchor position and regressing the bounding box directly. 
(d) Two-staged methods with an extra stage to correct the initial regression results.}  
\label{fig:Detection_anchor_roi}
\end{figure}

Later, researchers are drawing inspirations from the rapidly developing general object detection algorithms ~\citep{liu2016ssd,fu2017dssd,girshick2014rich,Girshick_2015_ICCV,ren2015faster,he2017mask}. 
In this stage, scene text detection algorithms are designed by modifying the region proposal and bounding box regression modules of general detectors to localize text instances directly~\citep{dai2017fused,He2017SSTD,jiang2017r2cnn,liao2017textboxes,liao2018textboxes++,Liu2017Deep,Shi_2017_CVPR,Yuliang2017Detecting,ma2017arbitrary,rong2017weakly,liao2018rotation,ZhangAAAI2018}, as shown in Fig. \ref{fig:Detection_anchor_roi}. 
They mainly consist of stacked convolutional layers that encode the input images into feature maps. 
Each spatial location at the feature map corresponds to a region of the input image. 
The feature maps are then fed into a classifier to predict the existence and localization of text instances at each such spatial location. 

These methods greatly reduce the pipeline into an end-to-end trainable neural network component, making training much easier and inference much faster. 
We introduce the most representative works here.  

Inspired by one-staged object detectors,  TextBoxes ~\citep{liao2017textboxes} adapts SSD ~\citep{liu2016ssd} to fit the varying orientations and aspect-ratios of text by defining default boxes as quadrilaterals with different aspect-ratio specs. 

EAST~\citep{Zhou_2017_CVPR} further simplifies the anchor-based detection by adopting the U-shaped design~\citep{Ronneberger2015U} to integrate features from different levels. 
Input images are encoded as one multi-channeled feature map instead of multiple layers of different spatial sizes in SSD. 
The feature at each spatial location is used to regress the rectangular or quadrilateral bounding box of the underlying text instances directly. 
Specifically, the existence of text, i.e. text/non-text, and geometries, e.g. orientation and size for rectangles, and vertexes coordinates for quadrilaterals, are predicted. EAST makes a difference to the field of text detection with its highly simplified pipeline and efficiency to perform inference at real-time speed.

Other methods adapt the two-staged object detection framework of R-CNN~\citep{girshick2014rich,Girshick_2015_ICCV,ren2015faster}, where the second stage corrects the localization results based on features obtained by Region of Interest (ROI) pooling. 

In ~\citep{ma2017arbitrary}, rotation region proposal networks are adapted to generate rotating region proposals, in order to fit into text of arbitrary orientations, instead of axis-aligned rectangles.

In FEN~\citep{ZhangAAAI2018}, the weighted sum of ROI poolings with different sizes is used. The final prediction is made by leveraging the \textit{textness} score for poolings of $4$ different sizes. 

\cite{zhang2019look} propose to perform ROI and localization branch recursively, to revise the predicted position of the text instance. It is a good way to include features at the boundaries of bounding boxes, which localizes the text better than region proposal networks (RPNs).

\cite{wang2018geometry}  propose to use a parametrized \textit{Instance Transformation Network} (ITN) that learns to predict appropriate affine transformation to perform on the last feature layer extracted by the base network, to rectify oriented text instances. Their method, with ITN, can be trained end-to-end. 

To adapt to irregularly shaped text, bounding polygons~\citep{Yuliang2017Detecting} with as many as $14$ vertexes are proposed, followed by a Bi-LSTM~\citep{hochreiter1997long} layer to refine the coordinates of the predicted vertexes. 

In a similar way, \cite{wang2019arbitrary} propose to use recurrent neural networks (RNNs) to read the features encoded by RPN-based two-staged object decoders and predict the bounding polygon with variable length. 
The method requires no post-processing or complex intermediate steps and achieves a much faster speed of $10.0$ FPS on Total-Text.

The main contribution in this stage is the simplification of the detection pipeline and the following improvement of efficiency. 
However, the performance is still limited when faced with curved, oriented, or long text for one-staged methods due to the limitation of the receptive field, and the efficiency is limited for two-staged methods. 

\subsubsection{Methods Based on Sub-Text Components}

The main distinction between text detection and general object detection is that text is homogeneous as a whole and is characterized by its locality, which is different from general object detection. 
By homogeneity and locality, we refer to the property that any part of a text instance is still text. 
Humans do not have to see the whole text instance to know it belongs to some text. 

\begin{figure}
\centering
\includegraphics[width=1.0\columnwidth]{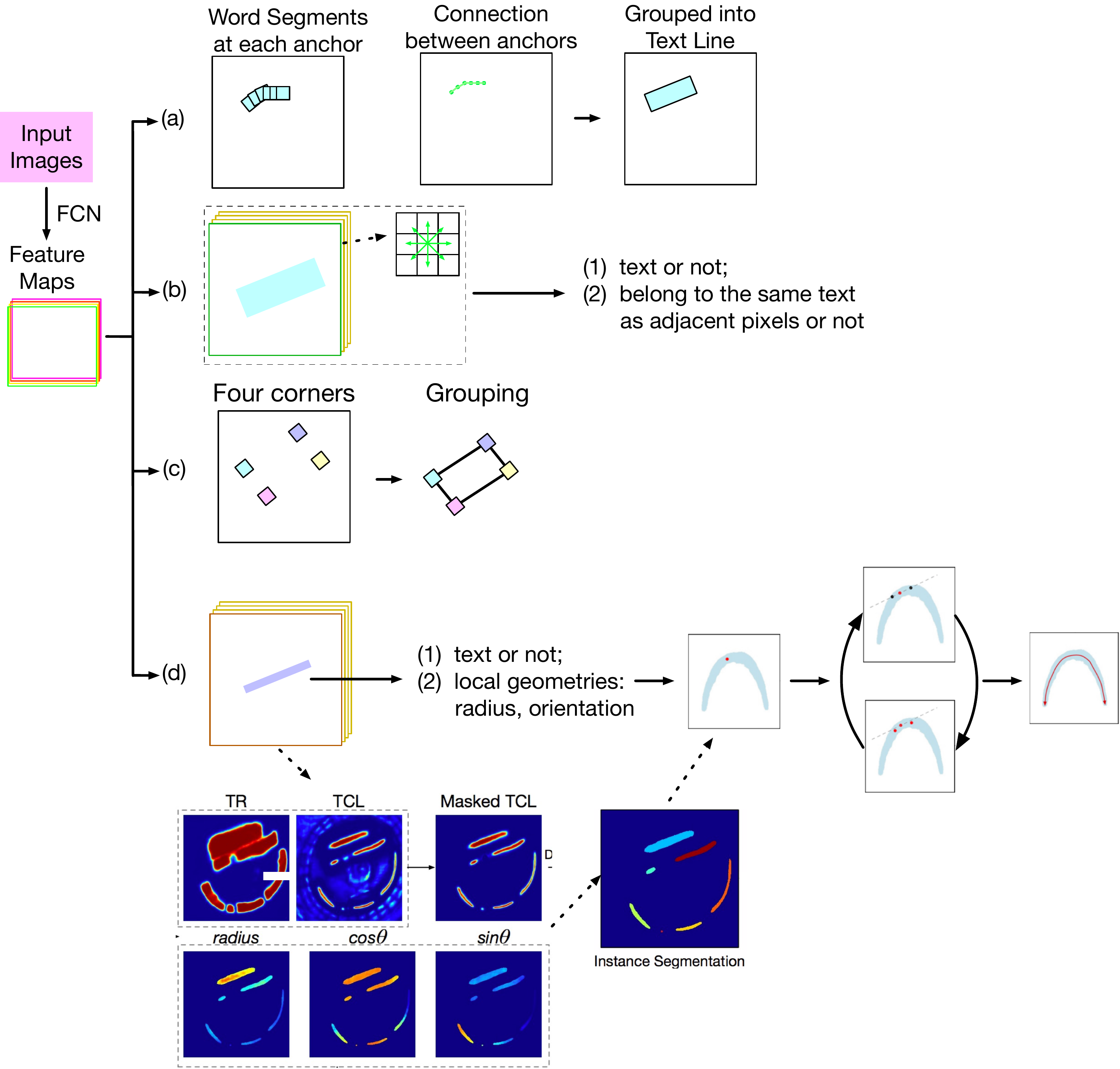}
\caption{Illustration of representative methods based on sub-text components: (a) SegLink~\citep{Shi_2017_CVPR}: with SSD as base network, predict word segments at each anchor position, and connections between adjacent anchors. (b) PixelLink~\citep{Deng2018}: for each pixel, predict text/non-text classification and whether it belongs to the same text as adjacent pixels or not. (c) Corner Localization~\citep{Lyu2018}: predict the four corners of each text and group those belonging to the same text instances. (d) TextSnake~\citep{long2018textsnake}: predict text/non-text and local geometries, which are used to reconstruct text instance. }
\label{fig:detection-bottom-up}
\end{figure}

Such a property lays a cornerstone for a new branch of text detection methods that only predict sub-text components and then assemble them into a text instance. 
These methods, by its nature, can better adapt to the aforementioned challenges of curved, long, and oriented text. 
These methods, as illustrated in Fig. \ref{fig:detection-bottom-up}, use neural networks to predict local attributes or segments, and a post-processing step to re-construct text instances. 
Compared with early multi-staged methods, they rely more on neural networks and have shorter pipelines.

In \textbf{pixel-level} methods~\citep{Deng2018,wu2017self}, an end-to-end fully convolutional neural network learns to generate a dense prediction map indicating whether each pixel in the original image belongs to any text instances or not. 
Post-processing methods then group pixels together depending on which pixels belong to the same text instance. 
Basically, they can be seen as a special case of instance segmentation~\citep{he2017mask}. 
Since text can appear in clusters which makes predicted pixels connected to each other,  the core of pixel-level methods is to separate text instances from each other. 

PixelLink~\citep{Deng2018} learns to predict whether two adjacent pixels belong to the same text instance by adding extra output channels to indicate links between adjacent pixels. 

Border learning method~\citep{wu2017self} casts each pixel into three categories: text, border, and background, assuming that the border can well separate text instances. 

In ~\citep{wang2017scene}, pixels are clustered according to their color consistency and edge information. 
The fused image segments are called \textit{superpixel}. These superpixels are further used to extract characters and predict text instances.

Upon the segmentation framework, \cite{tian2019learning} propose to add a loss term that maximizes the Euclidean distances between pixel embedding vectors that belong to different text instances, and minimizes those belonging to the same instance, to better separate adjacent texts.

\cite{wang2019shape} propose to predict text regions at different shrinkage scales, and enlarges the detected text region round-by-round, until collision with other instances. However, the prediction at different scales is itself a variation of the aforementioned border learning~\citep{wu2017self}.

\textbf{Component}-level methods usually predict at a medium granularity. Component refers to a local region of text instance, sometimes overlapping one or more characters. 

The representative component-level method is Connectionist Text Proposal Network (CTPN)~\citep{tian2016detecting}. 
CTPN models inherit the idea of anchoring and recurrent neural network for sequence labeling. 
They stack an RNN on top of CNNs. 
Each position in the final feature map represents features in the region specified by the corresponding anchor. 
Assuming that text appears horizontally, each row of features are fed into an RNN and labeled as text/non-text. 
Geometries such as segment sizes are also predicted. 
CTPN is the first to predict and connect segments of scene text with deep neural networks. 

SegLink~\citep{Shi_2017_CVPR} extends CTPN by considering the multi-oriented linkage between segments. 
The detection of segments is based on SSD~\citep{liu2016ssd}, where each default box represents a \textit{text segment}. 
Links between default boxes are predicted to indicate whether the adjacent segments belong to the same text instance.  \cite{zhang2020deep} further improve SegLink by using a Graph Convolutional Network~\citep{kipf2016semi} to predict the linkage between segments.  

Corner localization method~\citep{Lyu2018} proposes to detect the four corners of each text instance. 
Since each text instance only has $4$ corners, the prediction results and their relative position can indicate which corners should be grouped into the same text instance.

\begin{figure}
\centering
  \includegraphics[width=1.0\columnwidth]{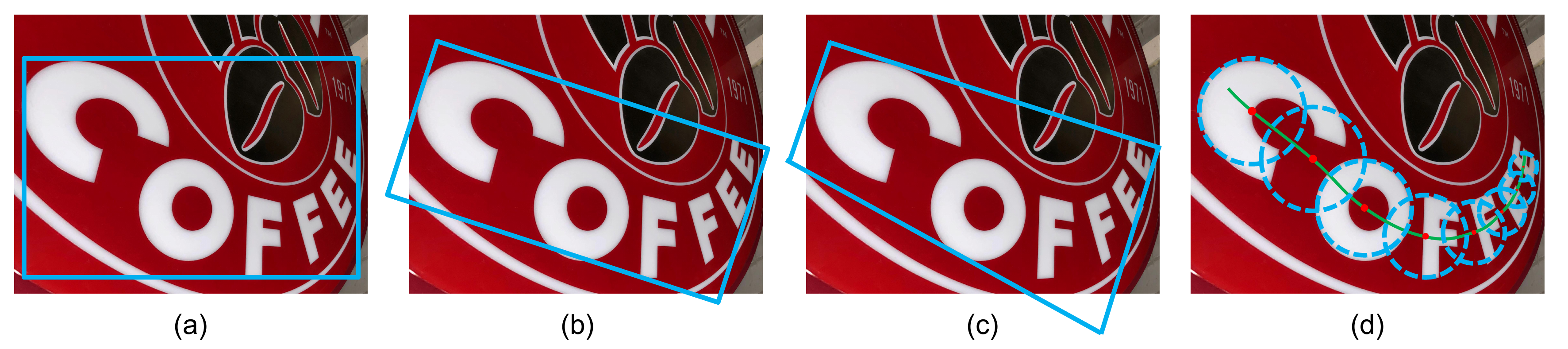}
\caption{{(a)-(c)}: Representing text as horizontal rectangles, oriented rectangles, and quadrilaterals. {(d)}: The sliding-disk representation proposed in TextSnake~\citep{long2018textsnake}.}
\label{fig:representation}     
\end{figure}

\cite{long2018textsnake} argue that text can be represented as a series of sliding round disks along the text center line (TCL), which is in accord with the running direction of the text instance, as shown in Fig.\ref{fig:representation}. With the novel representation, they present a new model, \textit{TextSnake}, which learns to predict local attributes, including TCL/non-TCL, text-region/non-text-region, radius, and orientation. The intersection of TCL pixels and text region pixels gives the final prediction of pixel-level TCL. Local geometries are then used to extract the TCL in the form of an ordered point list. With TCL and radius, the text line is reconstructed. It achieves state-of-the-art performance on several curved text datasets as well as more widely used ones, e.g. ICDAR 2015 \citep{karatzas2015icdar} and MSRA-TD 500~\citep{tu2012detecting}. Notably, Long \emph{et al.} propose a cross-validation test across different datasets, where models are only fine-tuned on datasets with straight text instances and tested on the curved datasets. In all existing curved datasets, TextSnake achieves improvements by up to $20\%$ over other baselines in F1-Score.

\textbf{Character-level} representation is yet another effective way.
\cite{baek2019character} propose to learn a segmentation map for character centers and links between them. 
Both components and links are predicted in the form of a Gaussian heat map. 
However, this method requires iterative weak supervision as real-world datasets are rarely equipped with character-level labels. 

Overall, detection based on sub-text components enjoys better flexibility and generalization ability over shapes and aspect ratios of text instance. 
The main drawback is that the module or post-processing step used to group segments into text instances may be vulnerable to noise, and the efficiency of this step is highly dependent on the actual implementation, and therefore may vary among different platforms.

\begin{figure*}
\centering
  \includegraphics[width=1.0\textwidth]{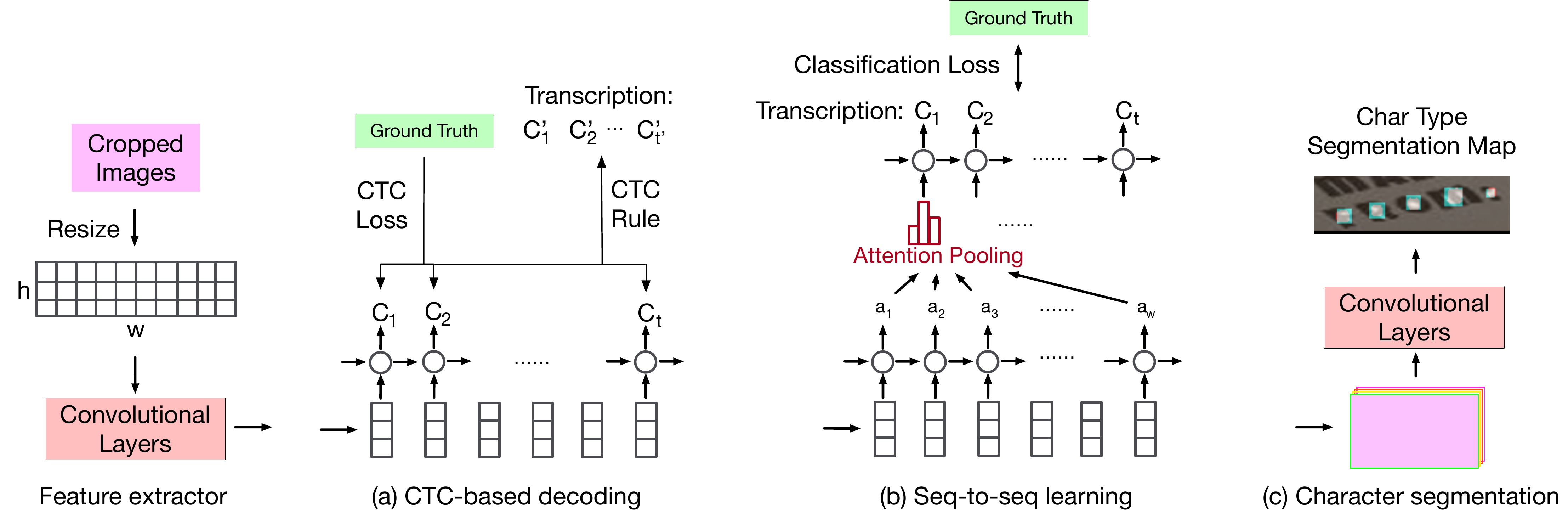}
\caption{Frameworks of text recognition models. 
(a) represents a sequence tagging model, and uses CTC for alignment in training and inference. 
(b) represents a sequence to sequence model, and can use cross-entropy to learn directly. 
(c) represents segmentation-based methods. 
}
\label{fig:recognition}      
\end{figure*}

\label{sec：4.2}
\subsection{Recognition}
In this section, we introduce methods for \textit{scene text recognition}. The input of these methods is cropped text instance images which contain only one word.

In the deep learning era, scene text recognition models use CNNs to encode images into feature spaces. 
The main difference lies in the text content decoding module. 
Two major techniques are the Connectionist Temporal Classification~\citep{graves2006connectionist} (CTC) and the encoder-decoder framework~\citep{sutskever2014sequence}. 
We introduce recognition methods in the literature based on the main technique they employ. 
Mainstream frameworks are illustrated in Fig.\ref{fig:recognition}.

Both CTC and encoder-decoder frameworks are originally designed for 1-dimensional sequential input data, and therefore are applicable to the recognition of straight and horizontal text, which can be encoded into a sequence of feature frames by CNNs without losing important information. 
However, characters in oriented and curved text are distributed over a 2-dimensional space. It remains a challenge to effectively represent oriented and curved text in feature spaces in order to fit the CTC and encoder-decoder frameworks, whose decodes require  1-dimensional inputs. For oriented and curved text, directly compressing the features into a 1-dimensional form may lose relevant information and bring in noise from background, thus leading to inferior recognition accuracy. 
We would introduce techniques to solve this challenge.

\subsubsection{CTC-Based Methods}
The CTC decoding module is adopted from speech recognition, where data are sequential in the time domain. 
To apply CTC in scene text recognition, the input images are viewed as a sequence of vertical pixel frames. 
The network outputs a per-frame prediction, indicating the probability distribution of label types for each frame. 
The CTC rule is then applied to edit the per-frame prediction to a text string. 
During training, the loss is computed as the sum of the negative log probability of all possible per-frame predictions that can generate the target sequence by CTC rules. 
Therefore, the CTC method makes it end-to-end trainable with only word-level annotations, without the need for character level annotations. 
The first application of CTC in the OCR domain can be traced to the handwriting recognition system of \cite{graves2008unconstrained}. 
Now this technique is widely adopted in scene text recognition~\citep{su2014accurate,he2016reading,liu2016star,gao2017reading,shi2017end,yin2017scene}.

The first attempts can be referred to as convolutional recurrent neural networks (CRNN). 
These models are composed by stacking RNNs on top of CNNs and use CTC for training and inference. 
DTRN~\citep{he2016reading} is the first CRNN model. 
It slides a CNN model across the input images to generate convolutional feature slices, which are then fed into RNNs. 
\citep{shi2017end} further improves DTRN by adopting the fully convolutional approach to encode the input images as a whole to generate features slices, utilizing the property that CNNs are not restricted by the spatial sizes of inputs. 

Instead of RNN, \cite{gao2017reading} adopt the stacked convolutional layers to effectively capture the contextual dependencies of the input sequence, which is characterized by lower computational complexity and easier parallel computation.

\cite{yin2017scene} simultaneously detect and recognize characters by sliding the text line image with character models, which are learned end-to-end on text line images labeled with text transcripts.

\subsubsection{Encoder-Decoder Methods}
The encoder-decoder framework for sequence-to-sequence learning is originally proposed in ~\citep{sutskever2014sequence} for machine translation. 
The encoder RNN reads an input sequence and passes its final latent state to a decoder RNN, which generates output in an auto-regressive way. 
The main advantage of the encoder-decoder framework is that it gives outputs of variable lengths, which satisfies the task setting of scene text recognition. 
The encoder-decoder framework is usually combined with the attention mechanism~\citep{bahdanau2014neural} which jointly learns to align input sequence and output sequence. 

\cite{lee2016recursive} present recursive recurrent neural networks with attention modeling for lexicon-free scene text recognition. the model first passes input images through recursive convolutional layers to extract encoded image features and then decodes them to output characters by recurrent neural networks with implicitly learned character-level language statistics. The attention-based mechanism performs soft feature selection for better image feature usage. 

\cite{cheng2017focusing} observe the attention drift problem in existing attention-based methods and proposes to impose localization supervision for attention score to attenuate it.

\cite{bai2018edit} propose an edit probability (EP) metric to handle the misalignment between the ground truth string and the attention's output sequence of the probability distribution. Unlike aforementioned attention-based methods, which usually employ a frame-wise maximal likelihood loss, EP tries to estimate the probability of generating a string from the output sequence of probability distribution conditioned on the input image, while considering the possible occurrences of missing or superfluous characters.

\cite{liu2018squeezedtext} propose an efficient attention-based encoder-decoder model, in which the encoder part is trained under binary constraints to reduce computation cost. 

Both CTC and the encoder-decoder framework simplify the recognition pipeline and make it possible to train scene text recognizers with only word-level annotations instead of character level annotations. 
Compared to CTC, the decoder module of the encoder-decoder framework is an implicit language model, and therefore, it can incorporate more linguistic priors. 
For the same reason, the encoder-decoder framework requires a larger training dataset with a larger vocabulary. 
Otherwise, the model may degenerate when reading words that are unseen during training. 
On the contrary, CTC is less dependent on language models and has a better character-to-pixel alignment. 
Therefore it is potentially better on languages such as Chinese and Japanese that have a large character set. 
The main drawback of these two methods is that they assume the text to be straight, and therefore can not adapt to irregular text. 

\subsubsection{Adaptions for Irregular Text Recognition}
Rectification-modules are a popular solution to irregular text recognition. 
\cite{shi2016robust,shi2018aster} propose a text recognition system which combined a Spatial Transformer Network (STN)~\citep{jaderberg2015spatial} and an attention-based Sequence Recognition Network. 
The STN-module predicts text bounding polygons with fully connected layers in order for Thin-Plate-Spline transformations which rectify the input irregular text image into a more canonical form, i.e. straight text. 
The rectification proves to be a successful strategy and forms the basis of the winning solution~\citep{long2019alchemy} in ICDAR 2019 ArT\footnote{\url{https://rrc.cvc.uab.es/?ch=14}} irregular text recognition competition. 

There have also been several improved versions of rectification based recognition. 
\cite{zhan2019esir} propose to perform rectification multiple times to gradually rectify the text. 
They also replace the text bounding polygons with a polynomial function to represent the shape. 
\cite{yang2019symmetry} propose to predict local attributes, such as radius and orientation values for pixels inside the text center region, in a similar way to TextSnake~\citep{long2018textsnake}. 
The orientation is defined as the orientation of the underlying character boxes, instead of text bounding polygons. 
Based on the attributes, bounding polygons are reconstructed in a way that the perspective distortion of characters is rectified, while the method by Shi \emph{et al.} and Zhan \emph{et al.} may only rectify at the text level and leave the characters distorted.

\cite{yang2017learning} introduce an auxiliary dense character detection task to encourage the learning of visual representations that are favorable to the text patterns. And they adopt an alignment loss to regularize the estimated attention at each time-step. Further, they use a coordinate map as a second input to enforce spatial-awareness. 

\cite{cheng2017arbitrarily} argue that encoding a text image as a 1-D sequence of features as implemented in most methods is not sufficient. They encode an input image to four feature sequences of four directions: horizontal, reversed horizontal, vertical, and reversed vertical. A weighting mechanism is applied to combine the four feature sequences.

\cite{liu2018char} present a hierarchical attention mechanism (HAM) which consists of a recurrent RoI-Warp layer and a character-level attention layer. 
They adopt a local transformation to model the distortion of individual characters, resulting in improved efficiency, and can handle different types of distortion that are hard to be modeled by a single global transformation. 

\cite{liao2018scene} cast the task of recognition into semantic segmentation, and treat each character type as one class. The method is insensitive to shapes and is thus effective on irregular text, but the lack of end-to-end training and sequence learning makes it prone to single-character errors, especially when the image quality is low. They are also the first to evaluate the robustness of their recognition method by padding and transforming test images. 

Another solution to irregular scene text recognition is 2-dimensional attention~\citep{xu2015show}, which has been verified in~\citep{li2018show}. 
Different from the sequential encoder-decoder framework, the 2D attentional model maintains 2-dimensional encoded features, and attention scores are computed for all spatial locations. Similar to spatial attention, \cite{long2020new} propose to first detect characters. 
Afterward, features are interpolated and gathered along the character center lines to form sequential feature frames.

In addition to the aforementioned techniques, \cite{qin2019towards} show that simply flattening the feature maps from 2-dimensional to 1-dimensional and feeding the resulting sequential features to RNN based attentional encoder-decoder model is sufficient to produce state-of-the-art recognition results on irregular text, which is a simple yet efficient solution. 

Apart from tailored model designs, \cite{long2019alchemy} synthesizes a curved text dataset, which significantly boosts the recognition performance on real-world curved text datasets with no sacrifices to straight text datasets. 

Although many elegant and neat solutions have been proposed, they are only evaluated and compared based on a relatively small dataset, CUTE80, which only consists of $288$ word samples. 
Besides, the training datasets used in these works only contain a negligible proportion of irregular text samples. 
Evaluations on larger datasets and more suitable training datasets may help us understand these methods better.

\subsubsection{Other Methods}

\cite{jaderberg2014deep,jaderberg2014synthetic} perform word recognition by classifying the image into a pre-defined set of vocabulary, under the framework of image classification. 
The model is trained by synthetic images, and achieves state-of-the-art performance on some benchmarks containing English words only. 
However, the application of this method is quite limited as it cannot be applied to recognize unseen sequences such as phone numbers and email addresses. 

To improve performance on difficult cases such as occlusion which brings ambiguity to single character recognition, \cite{yu2020towards} propose a transformer-based semantic reasoning module that performs translations from coarse, prone-to-error text outputs from the decoder to fine and linguistically calibrated outputs, which bears some resemblance to the deliberation networks for machine translation~\citep{xia2017deliberation} that first translate and then re-write the sentences. 

Despite the progress we have seen so far, the evaluation of recognition methods falls behind the time. As most detection methods can detect oriented and irregular text and some even rectify them, the recognition of such text may seem redundant. On the other hand, the robustness of recognition when cropped with a slightly different bounding box is seldom verified. Such robustness may be more important in real-world scenarios.

\label{sec：4.3}
\subsection{End-to-End System}

\begin{figure}
\centering
  \includegraphics[width=1.0\columnwidth]{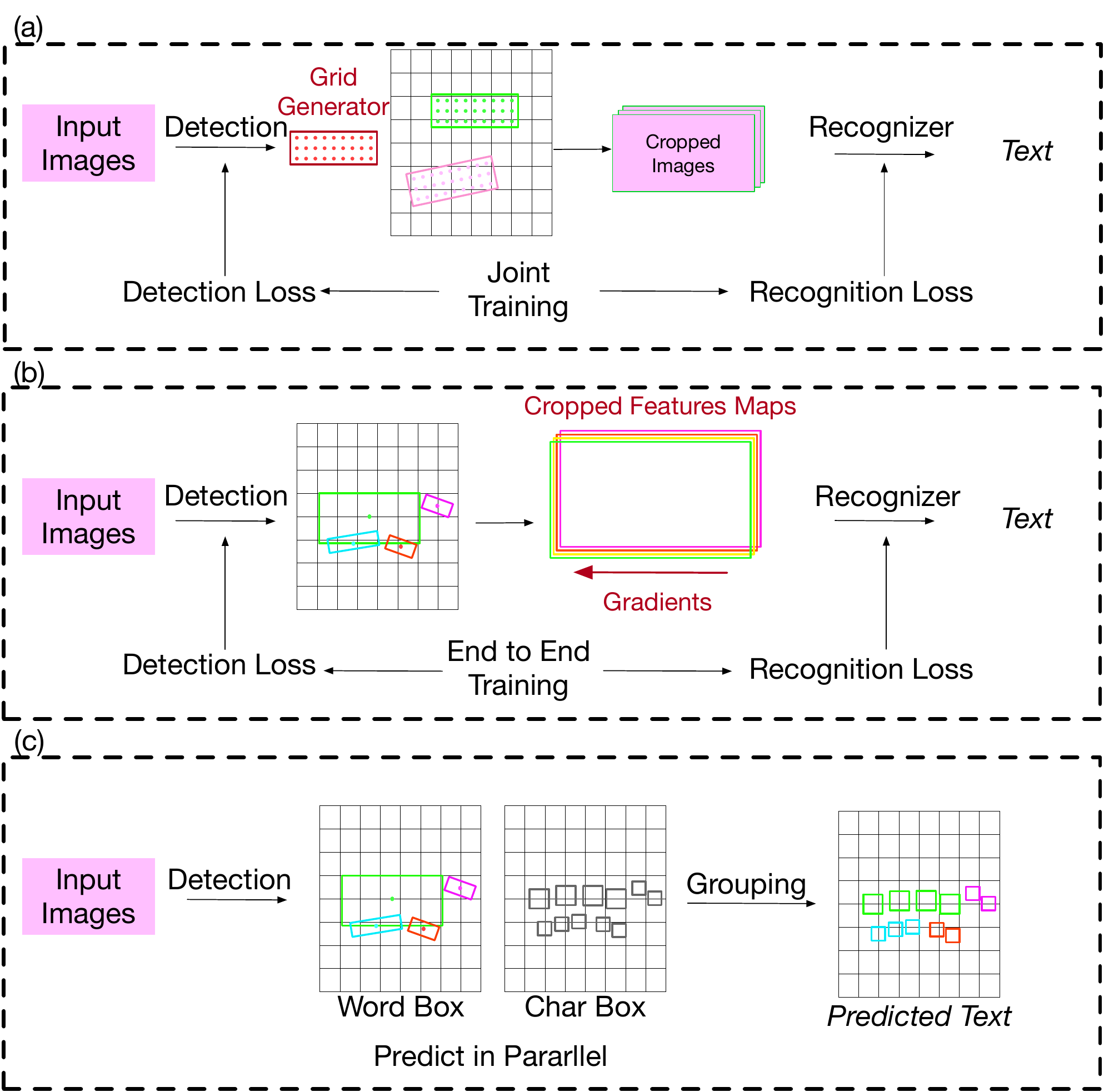}
\caption{Illustration of mainstream end-to-end frameworks. 
{(a)}: In SEE~\citep{bartz2017see}, the detection results are represented as grid matrices. Image regions are cropped and transformed before being fed into the recognition branch. 
{(b)}: Some methods crop from the feature maps and feed them to the recognition branch. 
{(c)}: While {(a)} and {(b)} utilize CTC-based and attention-based recognition branch, it is also possible to retrieve each character as generic objects and compose the text.}
\label{fig:end2end}       
\end{figure}

In the past, text detection and recognition are usually cast as two independent sub-problems that are combined to perform text reading from images. 
Recently, many end-to-end text detection and recognition systems (also known as text spotting systems) have been proposed, profiting a lot from the idea of designing differentiable computation graphs, as shown in Fig. \ref{fig:end2end}. 
Efforts to build such systems have gained considerable momentum as a new trend.

\noindent \textbf{Two-Step Pipelines} While earlier work~\citep{wang2011end,wang2012end} first detect single characters in the input image, recent systems usually detect and recognize text in word-level or line level.
Some of these systems first generate text proposals using a text detection model and then recognize them with another text recognition model~\citep{jaderberg2016reading,liao2017textboxes,gupta2016synthetic}. 
\cite{jaderberg2016reading} use a combination of Edge Box proposals~\citep{zitnick2014edge} and a trained aggregate channel features detector~\citep{dollar2014fast} to generate candidate word bounding boxes. 
Proposal boxes are filtered and rectified before being sent into their recognition model proposed in~\citep{jaderberg2014synthetic}.
~\cite{liao2017textboxes} combine an SSD~\citep{liu2016ssd} based text detector and CRNN~\citep{shi2017end} to spot text in images. 

In these methods, detected words are cropped from the image, and therefore, the detection and recognition are two separate steps. 
One major drawback of the two-step methods is that the propagation of error between the detection and recognition models will lead to less satisfactory performance. 

\noindent \textbf{Two-Stage Pipelines} Recently, end-to-end trainable networks are proposed to tackle this problem~\citep{bartz2017see,Busta_2017_ICCV,Li_2017_ICCV,he2018end,liu2018fots}, where feature maps instead of images are cropped and fed to recognition modules.

\cite{bartz2017see} present an solution that utilizes a STN~\citep{jaderberg2015spatial} to circularly attend to each word in the input image, and then recognize them separately. The united network is trained in a weakly-supervised manner that no word bounding box labels are used.
\cite{Li_2017_ICCV} substitute the object classification module in Faster-RCNN~\citep{ren2015faster} with an encoder-decoder based text recognition model and make up their text spotting system. 
\cite{liu2018fots}, ~\cite{Busta_2017_ICCV} and \cite{he2018end} develop unified text detection and recognition systems with a very similar overall architectures which consist of a detection branch and a recognition branch. 
\cite{liu2018fots} and \cite{Busta_2017_ICCV} adopt EAST~\citep{Zhou_2017_CVPR} and YOLOv2~\citep{redmon2017yolo9000} as their detection branches respectively, and have a similar text recognition branch in which text proposals are pooled into fixed height tensors by bilinear sampling and then transcribe into strings by a CTC-based recognition module. 
\cite{he2018end} also adopt EAST~\citep{Zhou_2017_CVPR} to generate text proposals, and they introduced character spatial information as explicit supervision in the attention-based recognition branch. 
\cite{lyu2018mask} propose a modification of Mask R-CNN. 
For each region of interest, character segmentation maps are produced, indicating the existence and location of a single character. 
A post-processing step that orders these character from left to right gives the final results.
In contrast to the aforementioned works that perform ROI pooling based on oriented bounding boxes, Qin \emph{et al.} ~\citep{qin2019towards} propose to use axis-aligned bounding boxes and mask the cropped features with a 0/1 textness segmentation mask ~\citep{he2017mask}. 

\noindent \textbf{One-Stage Pipeline} In addition to two-staged methods, \cite{xing2019convolutional} predict character and text bounding boxes as well as character type segmentation maps in parallel. 
The text bounding boxes are then used to group character boxes to form the final word transcription results. 
This is the first one-staged method.

\subsection{Auxiliary Techniques}

Recent advances are not limited to detection and recognition models that aim to solve the tasks directly. 
We should also give credit to auxiliary techniques that have played an important role.

\subsubsection{Synthetic Data}

Most deep learning models are data-thirsty. Their performance is guaranteed only when enough data are available. 
In the field of text detection and recognition, this problem is more urgent since most human-labeled datasets are small, usually containing around merely $1K-2K$ data instances. Fortunately, there have been work~\citep{jaderberg2014synthetic,gupta2016synthetic,zhan2018verisimilar,liao2019synthtext3d} that generate data of relatively high quality, and have been widely used for pre-training models for better performance. 

\cite{jaderberg2014synthetic} propose to generate synthetic data for text recognition. Their method blends text with randomly cropped natural images from human-labeled datasets after rending of font, border/shadow, color, and distortion. The results show that training merely on these synthetic data can achieve state-of-the-art performance and that synthetic data can act as augmentative data sources for all datasets. 

SynthText~\citep{gupta2016synthetic}
first propose to embed text in natural scene images for the training of text detection, while most previous work only print text on a cropped region and these synthetic data are only for text recognition. Printing text on the whole natural images poses new challenges, as it needs to maintain semantic coherence. To produce more realistic data, SynthText makes use of depth prediction~\citep{liu2015deep} and semantic segmentation~\citep{arbelaez2011contour}. Semantic segmentation groups pixels together as semantic clusters and each text instance is printed on one semantic surface, not overlapping multiple ones. A dense depth map is further used to determine the orientation and distortion of the text instance. The model trained only on SynthText achieves state-of-the-art on many text detection datasets. It is later used in other works~\citep{Zhou_2017_CVPR,Shi_2017_CVPR} as well for initial pre-training.

Further, \cite{zhan2018verisimilar} equip text synthesis with other deep learning techniques to produce more realistic samples. They introduce selective semantic segmentation so that word instances would only appear on sensible objects, e.g. a desk or wall in stead of someone's face. Text rendering in their work is adapted to the image so that they fit into the artistic styles and do not stand out awkwardly.

SynthText3D~\citep{liao2019synthtext3d} uses the famous open-source game engine, Unreal Engine 4 (UE4), and UnrealCV~\citep{qiu2017unrealcv} to synthesize scene text images. 
Text is rendered with the scene together and thus can achieve different lighting conditions, weather, and natural occlusions. 
However, SynthText3D simply follows the pipeline of SynthText and only makes use of the ground-truth depth and segmentation maps provided by the game engine. 
As a result, SynthText3D relies on manual selection of camera views, which limits its scalability. 
Besides, the proposed text regions are generated by clipping maximal rectangular bounding boxes extracted from segmentation maps, and therefore are limited to the middle parts of large and well-defined regions, which is an unfavorable location bias. 

UnrealText~\citep{long2020unrealtext} is another work using game engines to synthesize scene text images. 
It features deep interactions with the 3D worlds during synthesis. 
A ray-casting based algorithm is proposed to navigate in the 3D worlds efficiently and is able to generate diverse camera views automatically. 
The text region proposal module is based on collision detection and can put text onto the whole surfaces, thus getting rid of the location bias. 
UnrealText achieves significant speedup and better detector performances. 

\noindent \textbf{Text Editing} It is also worthwhile to mention the text editing task that is proposed recently~\citep{wu2019editing,yang2020swaptext}. 
Both works try to replace the text content while retaining text styles in natural images, such as the spatial arrangement of characters, text fonts, and colors. 
Text editing per se is useful in applications such as instant translation using cellphone cameras. 
It also has great potential in augmenting existing scene text images, though we have not seen any relevant experiment results yet.

\subsubsection{Weakly and Semi-Supervision}

\noindent \textbf{Bootstrapping for Character-Box}

\begin{figure}[H]
\centering
\includegraphics[width=1.0\columnwidth]{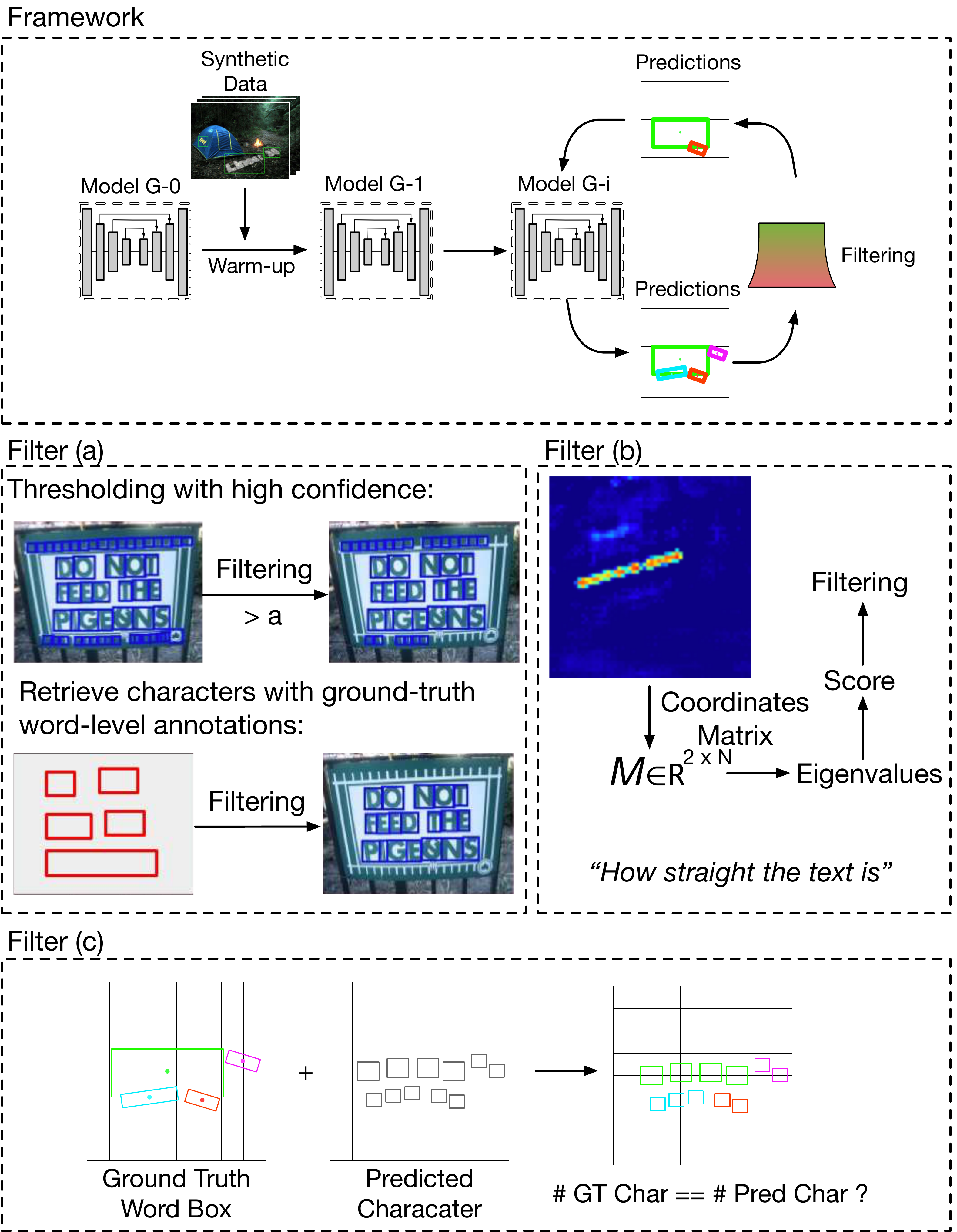}
\caption{Overview of semi-supervised and weakly-supervised methods. 
Existing methods differ in the way with regard to how filtering is done. 
{(a)}: WeText~\citep{tian2017wetext}, mainly by thresholding the confidence level and filtering by word-level annotation. 
{(b)}: Scoring-based methods, including WordSup~\citep{Hu2018wordsup} which assumes that text are straight lines, and uses a eigenvalue-based metric to measure its \textit{straightness}. 
{(c)}: by grouping characters into word using ground truth word bounding boxes, and comparing the number of characters~\citep{baek2019character,xing2019convolutional}. }  
\label{fig:bootstrapping}
\end{figure}

Character level annotations are more accurate and better. 
However, most existing datasets do not provide character-level annotating. 
Since characters are smaller and close to each other, character-level annotation is more costly and inconvenient. 
There has been some work on semi-supervised character detection. 
The basic idea is to initialize a character-detector and applies rules or threshold to pick the most reliable predicted candidates. 
These reliable candidates are then used as additional supervision sources to refine the character-detector. 
Both of them aim to augment existing datasets with character level annotations. 
Their difference is illustrated in Fig. \ref{fig:bootstrapping}.

WordSup~\citep{Hu2018wordsup} first initializes the character detector by training $5K$ warm-up iterations on synthetic datasets. 
For each image, WordSup generates character candidates, which are then filtered with word-boxes. For characters in each word box, the following score is computed to select the most possible character list:

\begin{equation}
\small
\centering
s = w\cdot \frac{area(B_{chars})}{area(B_{word})} + (1-w)\cdot(1-\frac{\lambda_2}{\lambda_1})
\label{eq-score-wetext}
\end{equation}

\noindent where $B_{chars}$ is the union of the selected character boxes; $B_{word}$ is the enclosing word bounding box; $\lambda_1$ and $\lambda_2$ are the first- and second-largest eigenvalues of a covariance matrix $C$, computed by the coordinates of the centers of the selected character boxes; $w$ is a weight scalar. Intuitively, the first term measures how complete the selected characters can cover the word boxes, while the second term measures whether the selected characters are located on a straight line, which is the main characteristic for word instances in most datasets.

WeText~\citep{tian2017wetext} start with a small dataset annotated on the character level. It follows two paradigms of bootstrapping: semi-supervised learning and weakly-supervised learning. In the semi-supervised setting, detected character candidates are filtered with a high thresholding value. In the weakly-supervised setting, ground-truth word boxes are used to mask out false positives outside. New instances detected in either way are added to the initial small datasets and re-train the model.

In ~\citep{baek2019character} and ~\citep{xing2019convolutional}, the character candidates are filtered with the help of word-level annotations. 
For each word instance, if the number of detected character bounding boxes inside the word bounding box equals to the length of the ground truth word, the character bounding boxes are regarded as correct. 

\noindent \textbf{Partial Annotations} 
In order to improve the recognition performance of end-to-end word spotting models on curved text, \cite{qin2019towards} propose to use off-the-shelf straight scene text spotting models to annotate a large number of unlabeled images. 
These images are called \textit{partially} labeled images, since the off-the-shelf models may omit some words. 
These partially annotated straight text prove to boost the performance on irregular text greatly.

Another similar effort is the large dataset proposed by ~\citep{sun2019chinese}, where each image is only annotated with one dominant text. 
They also design an algorithm to utilize these partially labeled data, which they claim are cheaper to annotate.

\begin{figure*}
\begin{centering}
  \includegraphics[width=2.0\columnwidth]{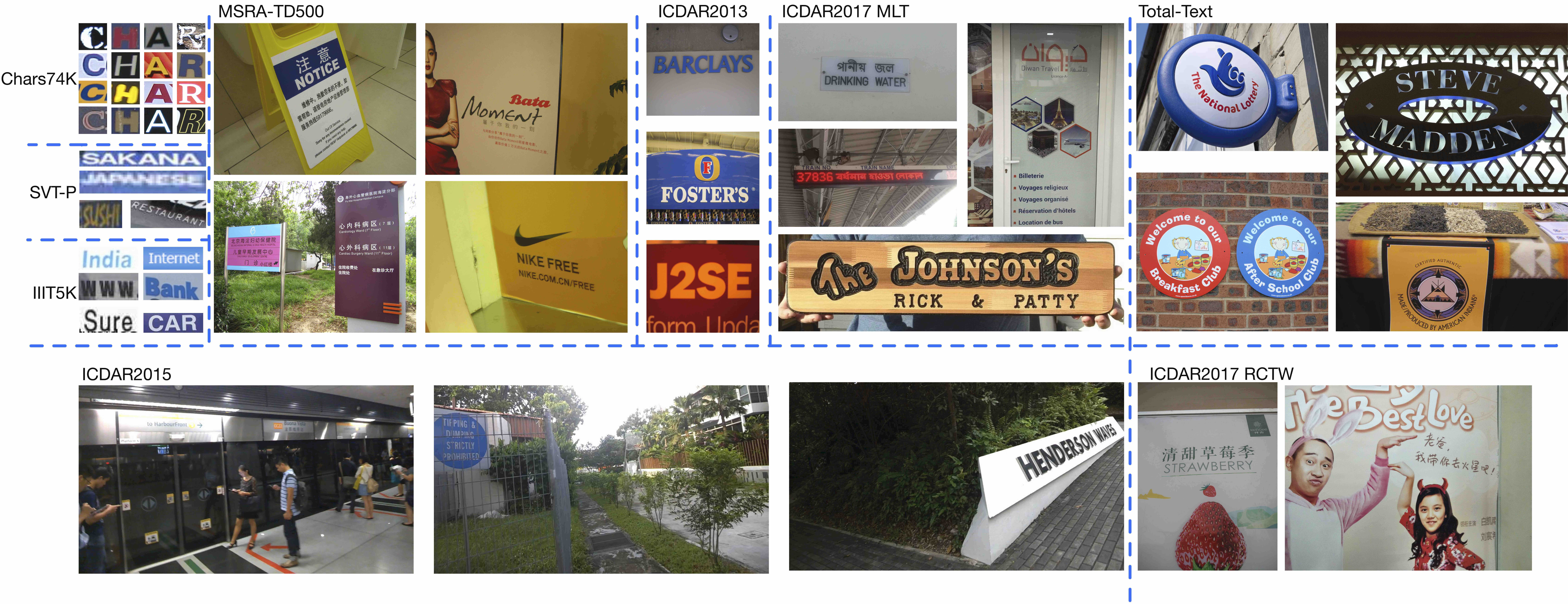}
\caption{Selected samples from \textit{Chars74K}, \textit{SVT-P}, \textit{IIIT5K}, \textit{MSRA-TD 500}, \textit{ICDAR 2013}, \textit{ICDAR 2015}, \textit{ICDAR 2017 MLT}, \textit{ICDAR 2017 RCTW}, and \textit{Total-Text}.}
\label{fig:datasets}
\par\end{centering}
\end{figure*}

\section{Benchmark Datasets and Evaluation Protocols}\label{sec-4}

As cutting edge algorithms achieved better performance on existing datasets, researchers are able to tackle more challenging aspects of the problems. New datasets aimed at different real-world challenges have been and are being crafted, benefiting the development of detection and recognition methods further.

In this section, we list and briefly introduce the existing datasets and the corresponding evaluation protocols. We also identify current state-of-the-art approaches to the widely used datasets when applicable.

\begin{table*}[!ht]
\centering
\caption{Public datasets for scene text detection and recognition. \textit{EN} stands for English and \textit{CN} stands for Chinese. 
Note that HUST-TR 400 is a supplementary training dataset for MSRA-TD 500. 
\textit{ICDAR 2013} refers to ICDAR 2013 Focused Scene Text Competition. 
\textit{ICDAR 2015} refers to ICDAR 2015 Incidental Text Competition. 
The last two columns indicate whether the datasets provide annotations for detection and recognition tasks. 
}
\label{tab-dataset}

\scalebox{0.95}{
\begin{tabular}{ccccccc}
\hline 
\textbf{\makecell{Dataset\\(Year)}} & \textbf{\makecell{Image Num\\(train/val/test)}} &  \textbf{\makecell{Orientation}}& \textbf{\makecell{Language}}& \textbf{\makecell{Features}}  &\textbf{Det.} & \textbf{Recog.} \tabularnewline
\hline
SVT (2010)& 100/0/250  & horizontal& EN& -  &  \checkmark  &  \checkmark  \tabularnewline
ICDAR 2003 & 258/0/251 & horizontal & EN & - & \checkmark & \checkmark\tabularnewline
ICDAR 2013
& 229/0/233 & horizontal & EN & stroke labels  &  \checkmark &  \checkmark  \tabularnewline
{CUTE} {(2014)}& 0/0/80 &  curved & EN& {-}  &  \checkmark  &  \checkmark\tabularnewline
ICDAR 2015
& 1000/0/500  & multi-oriented & EN & blur, small  &  \checkmark  &  \checkmark  \tabularnewline 
{ICDAR RCTW 2017}& 8034/0/4229 & multi-oriented & CN & -  &  \checkmark  &  \checkmark  \tabularnewline
{Total-Text} {(2017)}& 1255/0/300 & {curved} & EN, CN & polygon label  &  \checkmark  &  \checkmark   \tabularnewline
CTW (2017)& {25000/0/6000} & 
multi-oriented & CN & detailed attributes  &  \checkmark  &  \checkmark  \tabularnewline
{COCO-Text (2017)}& 43686/10000/10000 & 
multi-oriented & En & -  &  \checkmark  & \checkmark \tabularnewline
{ICDAR MLT 2017}& 7200/1800/9000 & 
multi-oriented & {9 langanges} & -  &  \checkmark  & \checkmark \tabularnewline
{ICDAR MLT 2019}& 10000/0/10000 & 
multi-oriented & {10 langanges} & -  &  \checkmark  & \checkmark \tabularnewline
{ICDAR ArT (2019)}& 5603/0/4563 & 
curved & EN, CN & -  &  \checkmark  & \checkmark \tabularnewline
{LSVT (2019)}& 20157/4968/4841 & 
multi-oriented & CN & 
{400K partially labeled images}
&  \checkmark  & \checkmark \tabularnewline
{MSRA-TD 500} {(2012)}& 300/0/200  &  multi-oriented & EN, CN & {long text}  &  \checkmark  & -  \tabularnewline
{HUST-TR 400} {(2014)}& 400/0/0 &  multi-oriented & EN, CN & {long text}  &  \checkmark  & - \tabularnewline
{CTW1500} {(2017)}& 1000/0/500  & 
{curved} & EN & -  
&  \checkmark  & - \tabularnewline
SVHN (2010)& 73257/0/26032 & horizontal& digits & household numbers   & - &  \checkmark  \tabularnewline
{IIIT 5K-Word} {(2012)}& 2000/0/3000 & horizontal & EN& -  & - &  \checkmark \tabularnewline
{SVTP} {(2013)} & 0/0/639  & multi-oriented & EN& {perspective text}  & - &  \checkmark \tabularnewline
\hline

\end{tabular} 
}
\end{table*}



\subsection{Benchmark Datasets}
We collect existing datasets and summarize their statistics in Tab.\ref{tab-dataset}. 
We select some representative image samples from some of the datasets, which are demonstrated in Fig.\ref{fig:datasets}. 
Links to these datasets are also collected in our Github repository mentioned in \textit{abstract}, for readers' convenience. 
In this section, we select some representative datasets and discuss their characteristics.

The ICDAR $2015$ incidental text focuses on small and oriented text. 
The images are taken by Google Glasses without taking care of the image quality. 
A large proportion of text in the images are very small,  blurred, occluded, and multi-oriented, making it very challenging.

The ICDAR MLT 2017 and 2019 datasets contain scripts of $9$ and $10$ languages respectively. 
They are the only multilingual datasets so far. 

Total-Text has a large proportion of curved text, while previous datasets contain only few. 
These images are mainly taken from street billboards, and annotated as polygons with a variable number of vertices.

\begin{table}
\small
\begin{centering}
\caption{Detection on ICDAR 2013.}  \label{tab_icdar2013} 
\begin{tabular}{llll}
\hline 
\textbf{Method} & \textbf{P} & \textbf{R} & \textbf{F1}  \tabularnewline
\hline 
\cite{zhang2016multi} & 88 & 78 & 83 \tabularnewline
\cite{gupta2016synthetic}& {92.0}  & {75.5} & {83.0} \tabularnewline
\cite{yao2016scene}& {88.88}  & {80.22} & {84.33}\tabularnewline
\cite{Deng2018}& {86.4}  & {83.6} & {84.5}  \tabularnewline
\cite{he2017multi}($*$) & 93 & 79  & 85  \tabularnewline
\cite{Shi_2017_CVPR} & 87.7  & 83.0 & 85.3  \tabularnewline
\cite{Lyu2018}& {93.3}  & {79.4} & {85.8} \tabularnewline
\cite{He_2017_ICCV} & 92  &  80 & 86 \tabularnewline
\cite{liao2017textboxes}& {89}  & {83} & {86} \tabularnewline
\cite{Zhou_2017_CVPR} & 92.64  & 82.67 & 87.37 \tabularnewline
\cite{liu2018learning}& {88.2}  & {87.2} & {87.7}  \tabularnewline
\cite{tian2016detecting} & 93  & 83  & 88  \tabularnewline
\cite{He2017SSTD}& {89}  & {86} & {88}  \tabularnewline
\cite{he2018end}& {88}  & {87} & {88} \tabularnewline
\cite{xue2018accurate}& {91.5}  & {87.1} & {89.2}  \tabularnewline
\cite{Hu2018wordsup}($*$) & 93.34  & 87.53  & 90.34  \tabularnewline
\cite{lyu2018mask}($*$) & {94.1}  & {88.1} & {91.0} \tabularnewline
\cite{ZhangAAAI2018}& {93.7}  & {90.0} & {92.3} \tabularnewline
\cite{baek2019character}& {97.4}  & {93.1} & {95.2} \tabularnewline\hline
\end{tabular}
\par\end{centering}
\end{table}

\begin{table}
\small
\begin{centering}
\caption{Detection on ICDAR MLT 2017.}  \label{tab_mlt17} 
\begin{tabular}{llll}
\hline 
\textbf{Method} & \textbf{P} & \textbf{R} & \textbf{F1} \tabularnewline
\hline 
\cite{liu2018fots} & 81.0 & 57.5 & 67.3  \tabularnewline
\cite{zhang2019look} & 60.6 & 78.8 & 68.5  \tabularnewline
\cite{wang2019shape} & 73.4 & 69.2 & 72.1 \tabularnewline
\cite{xing2019convolutional} & 70.10 & 77.07 & 73.42  \tabularnewline
\cite{baek2019character} & 68.2 & 80.6 & 73.9  \tabularnewline
\cite{long2020unrealtext} & 82.2 & 67.4 & 74.1 \tabularnewline
\hline 
\end{tabular}
\par\end{centering}
\end{table}

\begin{table}
\small
\begin{centering}
\caption{Detection on ICDAR 2015.} 
\label{tab_icdar2015}
\begin{tabular}{lllll}
\hline 
\textbf{Method} & \textbf{P} & \textbf{R} & \textbf{F1} & \textbf{FPS} \tabularnewline
\hline 
\cite{zhang2016multi} & 71 & 43.0 & 54 & 0.5 \tabularnewline
\cite{tian2016detecting} & 74  & 52  & 61 & - \tabularnewline
\cite{he2017multi}($*$) & 76  & 54  & 63 & - \tabularnewline
\cite{yao2016scene} & 72.26  & 58.69  & 64.77 & 1.6 \tabularnewline
\cite{Shi_2017_CVPR} & 73.1  & 76.8 & 75.0 & -   \tabularnewline
\cite{liu2018learning}& {72}  & {80} & {76} & -  \tabularnewline
\cite{He2017SSTD} & 80  & 73 & 77 & 7.7 \tabularnewline
\cite{Hu2018wordsup}($*$) & 79.33  & 77.03  & 78.16 & 2.0   \tabularnewline
\cite{Zhou_2017_CVPR} & 83.57  & 73.47 & 78.20 & 13.2  \tabularnewline
\cite{wang2018geometry}& {85.7}  & {74.1} & {79.5} & -   \tabularnewline
\cite{Lyu2018} & {94.1} & 70.7 & 80.7 & 3.6  \tabularnewline
\cite{He_2017_ICCV} & 82  &  80 & 81 & -  \tabularnewline
\cite{jiang2017r2cnn} & {85.62}  & {79.68} & {82.54} & -  \tabularnewline
{\cite{long2018textsnake}} & 84.9 & 80.4 & 82.6 & 10.2  \tabularnewline
\cite{he2018end}& {84}  & {83} & {83} & 1.1 \tabularnewline
{\cite{lyu2018mask}} & {85.8} & {81.2} & {83.4} & 4.8  \tabularnewline
\cite{Deng2018} & {85.5}  & {82.0} & {83.7} & 3.0  \tabularnewline
\cite{zhang2020deep} & {88.53}  & {84.69} & {86.56} & -  \tabularnewline
\cite{wang2019shape}& {86.92}  & {84.50} & {85.69} & 1.6 \tabularnewline
\cite{tian2019learning}& {88.3}  & {85.0} & {86.6} & 3   \tabularnewline
\cite{baek2019character}& {89.8}  & {84.3} & {86.9} & 8.6  \tabularnewline
\cite{zhang2019look}& {83.5}  & {91.3} & {87.2} & -  \tabularnewline
\cite{qin2019towards} & {89.36}  & {85.75} & {87.52} & 4.76  \tabularnewline
\cite{wang2019arbitrary}& {89.2}  & {86.0} & {87.6} & 10.0 \tabularnewline
\cite{xing2019convolutional} & {88.30}  & {91.15} & {89.70} & -  \tabularnewline\hline
\end{tabular}
\par\end{centering}
\end{table}

The Chinese Text in the Wild (CTW) dataset~\citep{yuan2018chinese} contains 
$32,285$ high-resolution street view images, annotated at the character level, including its underlying character type, bounding box, and detailed attributes such as whether it uses \textit{word-art}. 
The dataset is the largest one to date and the only one that contains detailed annotations. However, it only provides annotations for Chinese text and ignores other scripts, e.g. English.

LSVT~\citep{sun2019chinese} is composed of two datasets. 
One is fully labeled with word bounding boxes and word content. 
The other, while much larger, is only annotated with the word content of the dominant text instance. 
The authors propose to work on such partially labeled data that are much cheaper. 

\begin{table}
\small
\centering
\begin{centering}
\caption{Detection and end-to-end on Total-Text.}
\label{tab_total}
\begin{tabular}{lllll}
\hline 
\multirow{2}{*}{\textbf{Method}}  & \multicolumn{3}{l}{\textbf{\makecell{Detection}}} & \multirow{2}{*}{\textbf{E2E}} \tabularnewline
& \textbf{P} & \textbf{R} & \textbf{F} & \tabularnewline
\hline 
{\cite{lyu2018mask}} & {69.0} & {55.0} & {61.3} & 52.9   \tabularnewline
{\cite{long2018textsnake}} & {82.7} & {74.5} & {78.4}  & -  \tabularnewline

\cite{wang2019arbitrary}& {80.9}  & {76.2} & {78.5} & -\tabularnewline
\cite{wang2019shape}& {84.02}  & {77.96} & {80.87} & -\tabularnewline
\cite{zhang2019look}& {75.7}  & {88.6} & {81.6} & -\tabularnewline
\cite{baek2019character}& {87.6}  & {79.9} & {83.6}  & -\tabularnewline
\cite{qin2019towards}& {83.3}  & {83.4} & {83.3} & 67.8\tabularnewline
\cite{xing2019convolutional}& {81.0}  & {88.6} & {84.6} & 63.6\tabularnewline 
\cite{zhang2020deep}& {86.54}  & {84.93} & {85.73} & - \tabularnewline 
\hline
\end{tabular}
\par\end{centering}
\end{table}

\begin{table}
\small
\begin{centering}
\caption{Detection on CTW1500.} \label{tab_CTW1500}
\begin{tabular}{llll}
\hline 
\textbf{Method} & \textbf{P} & \textbf{R} & \textbf{F1}\tabularnewline
\hline 
\cite{Yuliang2017Detecting} & {77.4} & $69.8$ & $73.4$ \tabularnewline
{\cite{long2018textsnake}} & 67.9 & {85.3} & {75.6}  \tabularnewline
\cite{zhang2019look}& {69.6}  & {89.2} & {78.4} \tabularnewline
\cite{wang2019arbitrary}& {80.1}  & {80.2} & {80.1}\tabularnewline
\cite{tian2019learning}& {82.7}  & {77.8} & {80.1} \tabularnewline
\cite{wang2019shape} & 84.84 & 79.73 & 82.2 \tabularnewline
\cite{baek2019character}& {86.0}  & {81.1} & {83.5} \tabularnewline
\cite{zhang2020deep}& {85.93}  & {83.02} & {84.45} \tabularnewline
\hline
\end{tabular}
\par\end{centering}
\end{table}

IIIT 5K-Word~\citep{mishra2012scene} is the largest scene text recognition dataset, containing both digital and natural scene images. 
Its variance in font, color, size, and other noises makes it the most challenging one to date. 


\begin{table}
\small
\begin{centering}
\caption{Detection on MSRA-TD 500.}  \label{tab_td500}
\begin{tabular}{llll}
\hline 
\textbf{Method} & \textbf{P} & \textbf{R} & \textbf{F1} \tabularnewline
\hline 
\cite{kang2014orientation} & 71 &  62 & 66 \tabularnewline
\cite{zhang2016multi} & 83 & 67 & 74 \tabularnewline
\cite{He_2017_ICCV} & 77  &  70 & 74\tabularnewline
\cite{yao2016scene} & 76.51  & {75.31}  & 75.91 \tabularnewline
\cite{Zhou_2017_CVPR} & {87.28}  & {67.43} & 76.08  \tabularnewline
\cite{wu2017self} & {77}  & 78 & 77  \tabularnewline
\cite{Shi_2017_CVPR} & 86  & 70 & 77 \tabularnewline
\cite{Deng2018} & 83.0  & 73.2 & 77.8  \tabularnewline
{\cite{long2018textsnake}} & 83.2 & 73.9 & {78.3}  \tabularnewline
\cite{xue2018accurate}& {83.0}  & {77.4} & {80.1} \tabularnewline
\cite{wang2018geometry}& {90.3}  & {72.3} & {80.3} \tabularnewline
\cite{Lyu2018} & {87.6} & 76.2 & 81.5 \tabularnewline
\cite{baek2019character}& {88.2}  & {78.2} & {82.9}  \tabularnewline
\cite{tian2019learning}& {84.2}  & {81.7} & {82.9}  \tabularnewline
\cite{liu2018learning}& {88}  & {79} & {83} \tabularnewline
\cite{wang2019arbitrary}& {85.2}  & {82.1} & {83.6}  \tabularnewline
\cite{zhang2020deep}& {88.05}  & {82.30} & {85.08}  \tabularnewline
\hline
\end{tabular}
\par\end{centering}
\end{table}

\subsection{Evaluation Protocols}

In this part, we briefly summarize the evaluation protocols for text detection and recognition. 

As metrics for performance comparison of different algorithms, we usually refer to their precision, recall and F1-score. To compute these performance indicators, the list of predicted text instances should be matched to the ground truth labels in the first place. Precision, denoted as $P$, is calculated as the proportion of predicted text instances that can be matched to ground truth labels. Recall, denoted as $R$,  is the proportion of ground truth labels that have correspondents in the predicted list. F1-score is then computed by $F_1=\frac{2*P*R}{P+R}$, taking both precision and recall into account. Note that the matching between the predicted instances and ground truth ones comes first.

\subsubsection{Text Detection}
There are mainly two different protocols for text detection, the IOU based PASCAL Eval and overlap based DetEval. They differ in the criterion of matching predicted text instances and ground truth ones. In the following part, we use these notations: $S_{GT}$ is the area of the ground truth bounding box, $S_{P}$ is the area of the predicted bounding box, $S_{I}$ is the area of the intersection of the predicted and ground truth bounding box, $S_U$ is the area of the union.

\paragraph{$\bullet$ DetEval}: DetEval imposes constraints on both precision, i.e. $\frac{S_{I}}{S_{P}}$ and recall, i.e. $\frac{S_{I}}{S_{GT}}$. Only when both are larger than their respective thresholds, are they matched together.

\paragraph{$\bullet$ PASCAL}~\citep{everingham2015pascal}: The basic idea is that, if the intersection-over-union value, i.e. $\frac{S_{I}}{S_U}$, is larger than a designated threshold, the predicted and ground truth box are matched together.

Most works follow either one of the two evaluation protocols, but with small modifications. We only discuss those that are different from the two protocols mentioned above.

\paragraph{$\bullet$ ICDAR-2003/2005}: The match score $m$ is calculated in a way similar to IOU. It is defined as the ratio of the
area of intersection over that of the minimum bounding rectangular bounding box containing both.

\paragraph{$\bullet$ ICDAR-2011/2013}: One major drawback of the evaluation protocol of ICDAR2003/2005 is that it only considers the one-to-one match. It does not consider one-to-many, many-to-many, and many-to-one matching, which underestimates the actual performance. Therefore, ICDAR-2011/2013 follows the method proposed by ~\cite{wolf2006object}, where one-to-one matching is assigned a score of $1$ and the other two types are punished to a constant score less than $1$, usually set as $0.8$.

\begin{table}
\small
\begin{centering}
\caption{Characteristics of the three vocabulary lists used in ICDAR 2013/2015. \textit{S} stands for \textit{Strongly Contextualised}, \textit{W} for \textit{Weakly Contextualised}, and \textit{G} for \textit{Generic}}
\label{tab_icdar2015_vocab}
{
\begin{tabular}{cc}
\hline
\textbf{Vocab List} & \textbf{Description} \tabularnewline
\hline 
S & \makecell{per-image list of $100$ words\\ including all words in the image
}
\tabularnewline
W & all words in the entire test set
\tabularnewline
G & a $90k$-word generic vocabulary
\tabularnewline
\hline
\end{tabular}
}
\par\end{centering}
\end{table}

\paragraph{$\bullet$ MSRA-TD 500} \citep{tu2012detecting}: propose a new evaluation protocol for rotated bounding boxes, where both the predicted and ground truth bounding box are revolved horizontally around its center. They are matched only when the standard IOU score is higher than the threshold and the rotation of the original bounding boxes is less a pre-defined value (in practice $\frac{pi}{4}$).

\paragraph{$\bullet$ TIoU} \citep{liu2019tightness}: Tightness-IoU takes into account the fact that scene text recognition is sensitive to missing parts and superfluous parts in detection results. 
Not-retrieved areas will result in missing characters in recognition results, and redundant areas will result in unexpected characters. 
The proposed metrics penalize IoUs by scaling it down by the proportion of missing areas and the proportion of superfluous areas that overlap with other text.

The main drawback of existing evaluation protocols is that they only consider the best F1 scores under arbitrarily selected confidence thresholds selected on test sets. 
\cite{qin2019towards} also evaluate their method with the average precision (AP) metric that is widely adopted in general object detection. 
While F1 scores are only single points on the precision-recall curves, AP values consider the whole precision-recall curves. 
Therefore, AP is a more comprehensive metric and we urge that researchers in this field use AP instead of F1 alone.

\subsubsection{Text Recognition and End-to-End System}

In scene text recognition, the predicted text string is compared to the ground truth directly. 
The performance evaluation is in either character-level recognition rate (i.e. how many characters are recognized) or word level (whether the predicted word exactly the same as ground truth). 
ICDAR also introduces an edit-distance based performance evaluation.

\begin{table*}[h]
\caption{State-of-the-art recognition performance across a number of datasets. ``50'', ``1k'', ``Full'' are lexicons. ``0'' means no lexicon.  ``90k'' and ``ST'' are the Synth90k and the SynthText datasets, respectively. ``ST$^{+}$'' means including character-level annotations. ``Private'' means private training data.}
\label{sota-recognition}
\renewcommand{\arraystretch}{1.4}
\centering
\resizebox{\textwidth}{!}{%
\begin{tabular}{llccccccccccccc}
\hline 
\multirow{2}{*}{\textbf{Methods}} & \multirow{2}{*}{\textbf{ConvNet, Data}} & \multicolumn{3}{c}{\textbf{IIIT5k}} & \multicolumn{2}{c}{\textbf{SVT}} & \multicolumn{3}{c}{\textbf{IC03}} & \textbf{IC13} & \textbf{IC15} & \textbf{SVTP} & \textbf{CUTE} & \textbf{Total-Text}\tabularnewline
\cline{3-15} 
 &  & 50 & 1k & 0 & 50 & 0 & 50 & Full & 0 & 0 & 0 & 0 & 0 & 0\tabularnewline
\hline 
\cite{yao2014strokelets} & - & 80.2 & 69.3 & - & 75.9 & - & 88.5 & 80.3 & - & - & - & - & -&-\tabularnewline
\cite{jaderberg2014deepf} & - & - & - & - & 86.1 & - & 96.2 & 91.5 & - & - & - & - & -&-\tabularnewline
\cite{su2014accurate} & - & - & - & - & 83.0 & - & 92.0 & 82.0 & - & - & - & - & -&-\tabularnewline
\cite{gordo2015supervised} & - & 93.3 & 86.6 & - & 91.8 & - & - & - & - & - & - & - & -&-\tabularnewline
\cite{jaderberg2016reading} & VGG, 90k & 97.1 & 92.7 & - & 95.4 & 80.7 & 98.7 & {98.6} & 93.1 & 90.8 & - & - & -&-\tabularnewline
\cite{shi2017end} & VGG, 90k & 97.8 & 95.0 & 81.2 & 97.5 & 82.7 & 98.7 & 98.0 & 91.9 & 89.6 & - & - & -&-\tabularnewline
\cite{shi2016robust} & VGG, 90k & 96.2 & 93.8 & 81.9 & 95.5 & 81.9 & 98.3 & 96.2 & 90.1 & 88.6 & - & 71.8 & 59.2&-\tabularnewline
\cite{lee2016recursive} & VGG, 90k & 96.8 & 94.4 & 78.4 & 96.3 & 80.7 & 97.9 & 97.0 & 88.7 & 90.0 & - & - & -&-\tabularnewline
\cite{yang2017learning} & VGG, Private & 97.8 & 96.1 & - & 95.2 & - & 97.7 & - & - & - & - & 75.8 & 69.3&-\tabularnewline
\cite{cheng2017focusing} & ResNet, 90k + ST$^{+}$ & 99.3 & 97.5 & 87.4 & 97.1 & 85.9 & {99.2} & 97.3 & 94.2 & {93.3} & 70.6 & - & -&-\tabularnewline
\cite{shi2018aster} & ResNet, 90k + ST & {99.6} & {98.8} & {93.4} & {99.2} & {93.6} & 98.8 & 98.0 & {94.5} & 91.8 & {76.1} & {78.5} & {79.5}&-\tabularnewline
\cite{liao2018scene} & ResNet, ST$^{+}$ + Private & {99.8} & {98.8} & {91.9} & {98.8} & {86.4} & - & - & - & 91.5 & - & - & {79.9}&-\tabularnewline
\cite{li2018show} & ResNet, 90k + ST + Private & - & - & {91.5} & - & {84.5} & - & - & - & 91.0 & 69.2 & 76.4 & 83.3&-\tabularnewline
\cite{zhan2019esir} & ResNet, 90k + ST & 99.6 & 98.8 & {93.3} & 97.4 & {90.2} & - & - & - & 91.3 & 76.9 & 79.6 & 83.3 &-\tabularnewline
\cite{yang2019symmetry} & ResNet, 90k + ST & 99.5 & 98.8 & {94.4} & 97.2 & {88.9} & 99.0 & 98.3 & 95.0 & 93.9 & 78.7 & 80.8 & 87.5&- \tabularnewline
\cite{long2019alchemy} & ResNet, 90k + Curved ST & - & - & {94.8} & - & {89.6} & -& - & 95.8 & 92.8 & 78.2 & 81.6 & 89.6&76.3 \tabularnewline
\cite{yu2020towards} & ResNet, 90k + ST & - & - & {94.8} & - & {91.5} & -& - & - & 95.5 & 82.7 & 85.1 & 87.8&- \tabularnewline
\hline 
\end{tabular}
}
\label{tbl:comparison-to-sota}
\end{table*}

In end-to-end evaluation, matching is first performed in a similar way to that of text detection, and then the text content is compared. 

The most widely used datasets for end-to-end systems are ICDAR 2013~\citep{karatzas2013icdar} and ICDAR 2015~\citep{karatzas2015icdar}. 
The evaluation over these two datasets are carried out under two different settings\footnote{\url{http://rrc.cvc.uab.es/files/Robust\_Reading\_2015\_v02.pdf}}, 
the \textit{Word Spotting} setting and the \textit{End-to-End} setting. Under \textit{Word Spotting}, the performance evaluation only focuses on the text instances from the scene image that appears in a predesignated vocabulary, while other text instances are ignored. On the contrary, all text instances that appear in the scene image are included under \textit{End-to-End}. Three different vocabulary lists are provided for candidate transcriptions. They include \textit{Strongly Contextualised}, \textit{Weakly Contextualised}, and \textit{Generic}. The three kinds of lists are summarized in Tab.\ref{tab_icdar2015_vocab}. Note that under \textit{End-to-End}, these vocabularies can still serve as references.

Evaluation results of recent methods on several widely adopted benchmark datasets are summarized in the following tables: Tab. \ref{tab_icdar2013} for detection on ICDAR 2013, Tab. \ref{tab_icdar2015} for detection on ICDAR 2015 Incidental Text, Tab. \ref{tab_mlt17} for detection on ICDAR 2017 MLT, Tab. \ref{tab_total} for detection and end-to-end word spotting on Total-Text, Tab. \ref{tab_CTW1500} for detection on CTW1500, Tab. \ref{tab_td500} for detection on MSRA-TD 500, Tab. \ref{sota-recognition} for recognition on several datasets, and Tab. \ref{tab_icdar2015_e2e} for end-to-end text spotting on ICDAR 2013 and ICDAR 2015. 
Note that, we do not report performance under multi-scale conditions if single-scale performances are reported. 
We use $*$ to indicate methods where only multi-scale performances are reported. 
Since different backbone feature extractors are used in some works, we only report performances based on ResNet-50 unless not provided.

Note that, current evaluation for scene text recognition may be problematic. 
According to \cite{baek2019wrong}, most researchers are actually using different subsets when they refer to the same dataset, causing discrepancies in performance. 
Besides, \cite{long2020unrealtext} further point out that half of the widely adopted benchmark datasets have imperfect annotations, e.g. ignoring case-sensitivities and punctuations, and provide new annotations for those datasets. 
Though most paper claim to train their models to recognize in a case-sensitive way and also include punctuations, they may be limiting their output to only digits and case-insensitive characters during evaluation. 

\begin{table}
\small
\begin{centering}
\caption{Performance of End-to-End and Word Spotting on ICDAR 2015 and ICDAR 2013.}
\label{tab_icdar2015_e2e}
\scalebox{0.84}{
\begin{tabular}{lllllll}
\hline 
\multirow{2}{*}{\textbf{Method}}  & \multicolumn{3}{c}{\textbf{Word Spotting}} & \multicolumn{3}{c}{\textbf{End-to-End}}\tabularnewline
& \textbf{S} & \textbf{W} & \textbf{G} & \textbf{S} & \textbf{W} & \textbf{G} \tabularnewline
\hline 
\multicolumn{7}{c}{\textbf{ICDAR 2015}}\tabularnewline\hline 
\cite{liu2018fots}& 84.68 & 79.32 & 63.29 & 81.09 & 75.90 & 60.80 \tabularnewline
\cite{xing2019convolutional} 
& - & - & - & 80.14 & 74.45 & 62.18 \tabularnewline
\cite{lyu2018mask}& 79.3 & 74.5 & 64.2 & 79.3 & 73.0 & 62.4 \tabularnewline
\cite{he2018end}& 85 & 80 & 65 & 82 & 77 & 63 \tabularnewline
\cite{qin2019towards} 
& - & - & - & 83.38 & 79.94 & 67.98 \tabularnewline
\hline 
\multicolumn{7}{c}{\textbf{ICDAR 2013}}\tabularnewline\hline 
\cite{Busta_2017_ICCV}& 92 & 89 & 81 & 89 & 86 & 77 \tabularnewline
\cite{liu2018fots}& 92.73 & 90.72 & 83.51 & 88.81 & 87.11 & 80.81 \tabularnewline
\cite{Li_2017_ICCV}& 94.2 & 92.4 & 88.2 & 91.1 & 89.8 & 84.6 \tabularnewline
\cite{he2018end}& 93 & 92 & 87 & 91 & 89 & 86 \tabularnewline
\cite{lyu2018mask} & 92.5 & 92.0 & 88.2 & 92.2 & 91.1 & 86.5 \tabularnewline
\hline 
\end{tabular}
}
\par\end{centering}
\end{table}

\section{Application}\label{sec-5}

The detection and recognition of text---the visual and physical carrier of human civilization---allow the connection between vision and the understanding of its content further. Apart from the applications we have mentioned at the beginning of this paper, there have been numerous specific application scenarios across various industries and in our daily lives. In this part, we list and analyze the most outstanding ones that have, or are to have, significant impact, improving our productivity and life quality.

\noindent \textbf{\textsl{Automatic Data Entry}} Apart from an electronic archive of existing documents, OCR can also improve our productivity in the form of automatic data entry. Some industries involve time-consuming data type-in, e.g. express orders written by customers in the delivery industry, and hand-written information sheets in the financial and insurance industries. 
Applying OCR techniques can accelerate the data entry process as well as protect customer privacy. 
Some companies have already been using these technologies, e.g. SF-Express\footnote{Official website: \url{http://www.sf-express.com/cn/sc/}}. 
Another potential application is \textit{note taking}, such as NEBO\footnote{Official website: \url{https://www.myscript.com/nebo/}}, a note-taking software on tablets like iPad that performs instant transcription as users write down notes. 

\noindent \textbf{\textsl{Identity Authentication}}
Automatic identity authentication is yet another field where OCR can give a full play to. In fields such as Internet finance and Customs, users/passengers are required to provide identification (ID) information, such as identity card and passport. Automatic recognition and analysis of the provided documents would require OCR that reads and extracts the textual content, and can automate and greatly accelerate such processes. There are companies that have already started working on identification based on face and ID card, e.g. MEGVII (Face++)\footnote{\url{https://www.faceplusplus.com/face-based-identification/}}.

\noindent \textbf{\textsl{Augmented Computer Vision}}
As text is an essential element for the understanding of scene, OCR can assist computer vision in many ways. In the scenario of autonomous vehicles, text-embedded panels carry important information, e.g. geo-location, current traffic condition, navigation, and etc.. There have been several works on text detection and recognition for autonomous vehicle~\citep{mammeri2014road,mammeri2016mser}. The largest dataset so far, CTW~\citep{yuan2018chinese}, also places extra emphasis on traffic signs. Another example is the instant translation, where OCR is combined with a translation model. This is extremely helpful and time-saving as people travel or read documents written in foreign languages. Google's Translate application\footnote{\url{https://translate.google.com/}} can perform such instant translation. A similar application is instant text-to-speech software equipped with OCR, which can help those with visual disability and those who are illiterate\footnote{\url{https://en.wikipedia.org/wiki/Screen\_reader\#cite\_note-Braille\_display-2}}.

\noindent \textbf{\textsl{Intelligent Content Analysis}}
OCR also allows the industry to perform more intelligent analysis, mainly for platforms like video-sharing websites and e-commerce. Text can be extracted from images and subtitles as well as real-time commentary subtitles (a kind of floating comments added by users, e.g. those in Bilibili\footnote{\url{https://www.bilibili.com}} and Niconico\footnote{\url{www.nicovideo.jp/}}). On the one hand, such extracted text can be used in automatic content tagging and recommendation systems. They can also be used to perform user sentiment analysis, e.g. which part of the video attracts the users most. On the other hand, website administrators can impose supervision and filtration for inappropriate and illegal content, such as terrorist advocacy.

\section{Conclusion and Discussion}\label{sec-6}

\subsection{Status Quo}

\noindent \textbf{\textsl{Algorithms:}} The past several years have witnessed the significant development of algorithms for text detection and recognition, mainly due to the deep learning boom. 
Deep learning models have replaced the manual search and design for patterns and features. With the improved capability of models, research attention has been drawn to challenges such as oriented and curved text detection, and have achieved considerable progress.

\noindent \textbf{\textsl{Applications:}}
Apart from efforts towards a general solution to all sorts of images, these algorithms can be trained and adapted to more specific scenarios, e.g. \textit{bankcard}, \textit{ID card}, and \textit{driver's license}. Some companies have been providing such scenario-specific APIs, including Baidu Inc., Tencent Inc., and MEGVII Inc.. Recent development of fast and efficient methods~\citep{ren2015faster,Zhou_2017_CVPR} has also allowed the deployment of large-scale systems~\citep{borisyuk2018rosetta}. Companies including Google Inc. and Amazon Inc. are also providing text extraction APIs. 

\subsection{Challenges and Future Trends}

\textit{We look at the present through a rear-view mirror. We march backward into the future}~\citep{liu1975old}. We list and discuss challenges, and analyze what would be the next valuable research directions in the field scene text detection and recognition.

\noindent \textbf{\textsl{Languages:}} There are more than 1000 languages in the world\footnote{\url{https://www.ethnologue.com/guides/how-many-languages}}. However, most current algorithms and datasets have primarily focused on text of English. While English has a rather small alphabet, other languages such as Chinese and Japanese have a much larger one, with tens of thousands of symbols. RNN-based recognizers may suffer from such enlarged symbol sets. Moreover, some languages have much more complex appearances, and they are therefore more sensitive to conditions such as image quality. 
Researchers should first verify how well current algorithms can generalize to text of other languages and further to mixed text. Unified detection and recognition systems for multiple types of languages are of important academic value and application prospects. A feasible solution might be to explore compositional representations that can capture the common patterns of text instances of different languages, and train the detection and recognition models with text examples of different languages, which are generated by text synthesizing engines. 

\noindent \textbf{\textsl{Robustness of Models:}} Although current text recognizers have proven to be able to generalize well to different scene text datasets even only using synthetic data, recent work~\citep{liao2018scene} shows that robustness against flawed detection is not a neglectable problem. Actually, such instability in prediction has also been observed for text detection models. The reason behind this kind of phenomenon is still unclear. One conjecture is that the robustness of models is related to the internal operating mechanism of deep neural networks.

\noindent \textbf{\textsl{Generalization:}} Few detection algorithms except for TextSnake~\citep{long2018textsnake} have considered the problem of generalization ability across datasets, i.e. training on one dataset, and testing on another. Generalization ability is important as some application scenarios would require the adaptability to varying environments. For example, instant translation and OCR in autonomous vehicles should be able to perform stably under different situations: zoomed-in images with large text instances, far and small words, blurred words, different languages, and shapes. It remains unverified whether simply pooling all existing datasets together is enough, especially when the target domain is totally unknown. 

\noindent \textbf{\textsl{Evaluation:}}  
Existing evaluation metrics for detection stem from those for general object detection. Matching based on IoU score or pixel-level precision and recall ignore the fact that \textit{missing parts} and \textit{superfluous backgrounds} may hurt the performance of the subsequent recognition procedure. For each text instance, pixel-level precision and recall are good metrics. 
However, their scores are assigned to $1.0$ once they are matched to ground truth, and thus not reflected in the final dataset-level score. An off-the-shelf alternative method is to simply sum up the instance-level scores under DetEval instead of first assigning them to $1.0$. 

\noindent \textbf{\textsl{Synthetic Data:}}
While training recognizers on synthetic datasets has become a routine and results are excellent, detectors still rely heavily on real datasets. It remains a challenge to synthesize diverse and realistic images to train detectors. 
Potential benefits of synthetic data are not yet fully explored, such as generalization ability. 
Synthesis using 3D engines and models can simulate different conditions such as lighting and occlusion, and thus is worth further development.

\noindent \textbf{\textsl{Efficiency:}}
Another shortcoming of deep-learning-based methods lies in their efficiency. Most of the current systems can not run in real-time when deployed on computers without GPUs or mobile devices. Apart from model compression and lightweight models that have proven effective in other tasks, it is also valuable to study how to make custom speedup mechanism for text-related tasks.

\noindent \textbf{\textsl{Bigger and Better Datasets:}}
The sizes of most widely adopted datasets are small ($\sim 1k$ images). 
It will be worthwhile to study whether the improvements gained from current algorithms can scale up or they are just accidental results of better regularization. 
Besides, most datasets are only labeled with bounding boxes and texts. 
Detailed annotation of different attributes~\citep{yuan2018chinese} such as \textit{word-art} and \textit{occlusion} may guide researchers with pertinence. 
Finally, datasets characterized by real-world challenges are also important in advancing research progress, such as densely located text on products. 
Another related problem is that most of the existing datasets do not have validation sets. 
It is highly possible that the current reported evaluation results are actually upward biased due to overfitting on the test sets. 
We suggest that researchers should focus on large datasets, such as ICDAR MLT 2017, ICDAR MLT 2019, ICDAR ArT 2019, and COCO-Text.


\bibliography{Reference}   

\begin{thebibliography}{173}
\providecommand{\natexlab}[1]{#1}
\providecommand{\url}[1]{\texttt{#1}}
\expandafter\ifx\csname urlstyle\endcsname\relax
  \providecommand{\doi}[1]{doi: #1}\else
  \providecommand{\doi}{doi: \begingroup \urlstyle{rm}\Url}\fi

\bibitem[Almaz{\'a}n et~al.(2014)Almaz{\'a}n, Gordo, Forn{\'e}s, and
  Valveny]{almazan2014word}
J.~Almaz{\'a}n, A.~Gordo, A.~Forn{\'e}s, and E.~Valveny.
\newblock Word spotting and recognition with embedded attributes.
\newblock \emph{IEEE transactions on pattern analysis and machine
  intelligence}, 36\penalty0 (12):\penalty0 2552--2566, 2014.

\bibitem[Arbelaez et~al.(2011)Arbelaez, Maire, Fowlkes, and
  Malik]{arbelaez2011contour}
P.~Arbelaez, M.~Maire, C.~Fowlkes, and J.~Malik.
\newblock Contour detection and hierarchical image segmentation.
\newblock \emph{IEEE transactions on pattern analysis and machine
  intelligence}, 33\penalty0 (5):\penalty0 898--916, 2011.

\bibitem[Baek et~al.(2019{\natexlab{a}})Baek, Kim, Lee, Park, Han, Yun, Oh, and
  Lee]{baek2019wrong}
J.~Baek, G.~Kim, J.~Lee, S.~Park, D.~Han, S.~Yun, S.~J. Oh, and H.~Lee.
\newblock What is wrong with scene text recognition model comparisons? dataset
  and model analysis.
\newblock In \emph{Proceedings of the IEEE International Conference on Computer
  Vision}, pages 4715--4723, 2019{\natexlab{a}}.

\bibitem[Baek et~al.(2019{\natexlab{b}})Baek, Lee, Han, Yun, and
  Lee]{baek2019character}
Y.~Baek, B.~Lee, D.~Han, S.~Yun, and H.~Lee.
\newblock Character region awareness for text detection.
\newblock In \emph{Proceedings of the IEEE Conference on Computer Vision and
  Pattern Recognition (CVPR)}, pages 9365--9374, 2019{\natexlab{b}}.

\bibitem[Bahdanau et~al.(2014)Bahdanau, Cho, and Bengio]{bahdanau2014neural}
D.~Bahdanau, K.~Cho, and Y.~Bengio.
\newblock Neural machine translation by jointly learning to align and
  translate.
\newblock \emph{ICLR 2015}, 2014.

\bibitem[Bai et~al.(2018)Bai, Cheng, Niu, Pu, and Zhou]{bai2018edit}
F.~Bai, Z.~Cheng, Y.~Niu, S.~Pu, and S.~Zhou.
\newblock Edit probability for scene text recognition.
\newblock In \emph{CVPR 2018}, 2018.

\bibitem[Bartz et~al.(2017)Bartz, Yang, and Meinel]{bartz2017see}
C.~Bartz, H.~Yang, and C.~Meinel.
\newblock See: Towards semi-supervised end-to-end scene text recognition.
\newblock \emph{arXiv preprint arXiv:1712.05404}, 2017.

\bibitem[Bissacco et~al.(2013)Bissacco, Cummins, Netzer, and
  Neven]{bissacco2013photoocr}
A.~Bissacco, M.~Cummins, Y.~Netzer, and H.~Neven.
\newblock Photoocr: Reading text in uncontrolled conditions.
\newblock In \emph{Proceedings of the IEEE International Conference on Computer
  Vision}, pages 785--792, 2013.

\bibitem[Borisyuk et~al.(2018)Borisyuk, Gordo, and
  Sivakumar]{borisyuk2018rosetta}
F.~Borisyuk, A.~Gordo, and V.~Sivakumar.
\newblock Rosetta: Large scale system for text detection and recognition in
  images.
\newblock In \emph{Proceedings of the 24th ACM SIGKDD International Conference
  on Knowledge Discovery \& Data Mining}, pages 71--79. ACM, 2018.

\bibitem[Busta et~al.(2015)Busta, Neumann, and Matas]{busta2015fastext}
M.~Busta, L.~Neumann, and J.~Matas.
\newblock Fastext: Efficient unconstrained scene text detector.
\newblock In \emph{Proceedings of the IEEE International Conference on Computer
  Vision (ICCV)}, pages 1206--1214, 2015.

\bibitem[Busta et~al.(2017)Busta, Neumann, and Matas]{Busta_2017_ICCV}
M.~Busta, L.~Neumann, and J.~Matas.
\newblock Deep textspotter: An end-to-end trainable scene text localization and
  recognition framework.
\newblock In \emph{Proc. ICCV}, 2017.

\bibitem[Chen et~al.(2004)Chen, Yang, Zhang, and Waibel]{chen2004automatic}
X.~Chen, J.~Yang, J.~Zhang, and A.~Waibel.
\newblock Automatic detection and recognition of signs from natural scenes.
\newblock \emph{IEEE Transactions on image processing}, 13\penalty0
  (1):\penalty0 87--99, 2004.

\bibitem[Cheng et~al.(2017{\natexlab{a}})Cheng, Bai, Xu, Zheng, Pu, and
  Zhou]{cheng2017focusing}
Z.~Cheng, F.~Bai, Y.~Xu, G.~Zheng, S.~Pu, and S.~Zhou.
\newblock Focusing attention: Towards accurate text recognition in natural
  images.
\newblock In \emph{2017 IEEE International Conference on Computer Vision
  (ICCV)}, pages 5086--5094. IEEE, 2017{\natexlab{a}}.

\bibitem[Cheng et~al.(2017{\natexlab{b}})Cheng, Liu, Bai, Niu, Pu, and
  Zhou]{cheng2017arbitrarily}
Z.~Cheng, X.~Liu, F.~Bai, Y.~Niu, S.~Pu, and S.~Zhou.
\newblock Arbitrarily-oriented text recognition.
\newblock \emph{CVPR2018}, 2017{\natexlab{b}}.

\bibitem[Ch'ng and Chan(2017)]{kheng2017total}
C.~K. Ch'ng and C.~S. Chan.
\newblock Total-text: A comprehensive dataset for scene text detection and
  recognition.
\newblock In \emph{2017 14th IAPR International Conference on Document Analysis
  and Recognition (ICDAR)}, volume~1, pages 935--942. IEEE, 2017.

\bibitem[Chowdhury and Deb(2013)]{chowdhury2013extracting}
M.~A. Chowdhury and K.~Deb.
\newblock Extracting and segmenting container name from container images.
\newblock \emph{International Journal of Computer Applications}, 74\penalty0
  (19), 2013.

\bibitem[Coates et~al.(2011)Coates, Carpenter, Case, Satheesh, Suresh, Wang,
  Wu, and Ng]{coates2011text}
A.~Coates, B.~Carpenter, C.~Case, S.~Satheesh, B.~Suresh, T.~Wang, D.~J. Wu,
  and A.~Y. Ng.
\newblock Text detection and character recognition in scene images with
  unsupervised feature learning.
\newblock In \emph{2011 International Conference on Document Analysis and
  Recognition (ICDAR)}, pages 440--445. IEEE, 2011.

\bibitem[Dai et~al.(2017)Dai, Huang, Gao, and Chen]{dai2017fused}
Y.~Dai, Z.~Huang, Y.~Gao, and K.~Chen.
\newblock Fused text segmentation networks for multi-oriented scene text
  detection.
\newblock \emph{arXiv preprint arXiv:1709.03272}, 2017.

\bibitem[Dalal and Triggs(2005)]{dalal2005histograms}
N.~Dalal and B.~Triggs.
\newblock Histograms of oriented gradients for human detection.
\newblock In \emph{IEEE Computer Society Conference on Computer Vision and
  Pattern Recognition (CVPR)}, volume~1, pages 886--893. IEEE, 2005.

\bibitem[Deng et~al.(2018)Deng, Liu, Li, and Cai]{Deng2018}
D.~Deng, H.~Liu, X.~Li, and D.~Cai.
\newblock Pixellink: Detecting scene text via instance segmentation.
\newblock In \emph{Proceedings of AAAI, 2018}, 2018.

\bibitem[DeSouza and Kak(2002)]{desouza2002vision}
G.~N. DeSouza and A.~C. Kak.
\newblock Vision for mobile robot navigation: A survey.
\newblock \emph{IEEE transactions on pattern analysis and machine
  intelligence}, 24\penalty0 (2):\penalty0 237--267, 2002.

\bibitem[Doll{\'a}r et~al.(2014)Doll{\'a}r, Appel, Belongie, and
  Perona]{dollar2014fast}
P.~Doll{\'a}r, R.~Appel, S.~Belongie, and P.~Perona.
\newblock Fast feature pyramids for object detection.
\newblock \emph{IEEE Transactions on Pattern Analysis and Machine
  Intelligence}, 36\penalty0 (8):\penalty0 1532--1545, 2014.

\bibitem[Dvorin and Havosha(2009)]{dvorin2009method}
Y.~Dvorin and U.~E. Havosha.
\newblock Method and device for instant translation, June~4 2009.
\newblock US Patent App. 11/998,931.

\bibitem[Epshtein et~al.(2010)Epshtein, Ofek, and
  Wexler]{epshtein2010detecting}
B.~Epshtein, E.~Ofek, and Y.~Wexler.
\newblock Detecting text in natural scenes with stroke width transform.
\newblock In \emph{2010 IEEE Conference on Computer Vision and Pattern
  Recognition (CVPR)}, pages 2963--2970. IEEE, 2010.

\bibitem[Everingham et~al.(2015)Everingham, Eslami, Van~Gool, Williams, Winn,
  and Zisserman]{everingham2015pascal}
M.~Everingham, S.~A. Eslami, L.~Van~Gool, C.~K. Williams, J.~Winn, and
  A.~Zisserman.
\newblock The pascal visual object classes challenge: A retrospective.
\newblock \emph{International journal of computer vision}, 111\penalty0
  (1):\penalty0 98--136, 2015.

\bibitem[Felzenszwalb and Huttenlocher(2005)]{felzenszwalb2005pictorial}
P.~F. Felzenszwalb and D.~P. Huttenlocher.
\newblock Pictorial structures for object recognition.
\newblock \emph{International journal of computer vision}, 61\penalty0
  (1):\penalty0 55--79, 2005.

\bibitem[Fu et~al.(2017)Fu, Liu, Ranga, Tyagi, and Berg]{fu2017dssd}
C.-Y. Fu, W.~Liu, A.~Ranga, A.~Tyagi, and A.~C. Berg.
\newblock Dssd: Deconvolutional single shot detector.
\newblock \emph{arXiv preprint arXiv:1701.06659}, 2017.

\bibitem[Gao et~al.(2017)Gao, Chen, Wang, and Lu]{gao2017reading}
Y.~Gao, Y.~Chen, J.~Wang, and H.~Lu.
\newblock Reading scene text with attention convolutional sequence modeling.
\newblock \emph{arXiv preprint arXiv:1709.04303}, 2017.

\bibitem[Girshick(2015)]{Girshick_2015_ICCV}
R.~Girshick.
\newblock Fast r-cnn.
\newblock In \emph{The IEEE International Conference on Computer Vision
  (ICCV)}, 2015.

\bibitem[Girshick et~al.(2014)Girshick, Donahue, Darrell, and
  Malik]{girshick2014rich}
R.~Girshick, J.~Donahue, T.~Darrell, and J.~Malik.
\newblock Rich feature hierarchies for accurate object detection and semantic
  segmentation.
\newblock In \emph{Proceedings of the IEEE conference on computer vision and
  pattern recognition (CVPR)}, pages 580--587, 2014.

\bibitem[Goldberg(1997)]{goldberg1997efficient}
A.~V. Goldberg.
\newblock An efficient implementation of a scaling minimum-cost flow algorithm.
\newblock \emph{Journal of algorithms}, 22\penalty0 (1):\penalty0 1--29, 1997.

\bibitem[Gordo(2015)]{gordo2015supervised}
A.~Gordo.
\newblock Supervised mid-level features for word image representation.
\newblock In \emph{Proceedings of the IEEE conference on computer vision and
  pattern recognition (CVPR)}, pages 2956--2964, 2015.

\bibitem[Graves et~al.(2006)Graves, Fern{\'a}ndez, Gomez, and
  Schmidhuber]{graves2006connectionist}
A.~Graves, S.~Fern{\'a}ndez, F.~Gomez, and J.~Schmidhuber.
\newblock Connectionist temporal classification: labelling unsegmented sequence
  data with recurrent neural networks.
\newblock In \emph{Proceedings of the 23rd international conference on Machine
  learning}, pages 369--376. ACM, 2006.

\bibitem[Graves et~al.(2008)Graves, Liwicki, Bunke, Schmidhuber, and
  Fern{\'a}ndez]{graves2008unconstrained}
A.~Graves, M.~Liwicki, H.~Bunke, J.~Schmidhuber, and S.~Fern{\'a}ndez.
\newblock Unconstrained on-line handwriting recognition with recurrent neural
  networks.
\newblock In \emph{Advances in neural information processing systems}, pages
  577--584, 2008.

\bibitem[Gupta et~al.(2016)Gupta, Vedaldi, and Zisserman]{gupta2016synthetic}
A.~Gupta, A.~Vedaldi, and A.~Zisserman.
\newblock Synthetic data for text localisation in natural images.
\newblock In \emph{Proceedings of the IEEE Conference on Computer Vision and
  Pattern Recognition (CVPR)}, pages 2315--2324, 2016.

\bibitem[Ham et~al.(1995)Ham, Kang, Chung, Park, and Park]{ham1995recognition}
Y.~K. Ham, M.~S. Kang, H.~K. Chung, R.-H. Park, and G.~T. Park.
\newblock Recognition of raised characters for automatic classification of
  rubber tires.
\newblock \emph{Optical Engineering}, 34\penalty0 (1):\penalty0 102--110, 1995.

\bibitem[Han et~al.(2018)Han, Zhang, Cheng, Liu, and Xu]{han2018advanced}
J.~Han, D.~Zhang, G.~Cheng, N.~Liu, and D.~Xu.
\newblock Advanced deep-learning techniques for salient and category-specific
  object detection: a survey.
\newblock \emph{IEEE Signal Processing Magazine}, 35\penalty0 (1):\penalty0
  84--100, 2018.

\bibitem[He et~al.(2017{\natexlab{a}})He, Yang, Liang, Zhou, Ororbia, Kifer,
  and Giles]{he2017multi}
D.~He, X.~Yang, C.~Liang, Z.~Zhou, A.~G. Ororbia, D.~Kifer, and C.~L. Giles.
\newblock Multi-scale fcn with cascaded instance aware segmentation for
  arbitrary oriented word spotting in the wild.
\newblock In \emph{2017 IEEE Conference on Computer Vision and Pattern
  Recognition (CVPR)}, pages 474--483. IEEE, 2017{\natexlab{a}}.

\bibitem[He et~al.(2017{\natexlab{b}})He, Gkioxari, Doll{\'a}r, and
  Girshick]{he2017mask}
K.~He, G.~Gkioxari, P.~Doll{\'a}r, and R.~Girshick.
\newblock Mask r-cnn.
\newblock In \emph{2017 IEEE International Conference on Computer Vision
  (ICCV)}, pages 2980--2988. IEEE, 2017{\natexlab{b}}.

\bibitem[He et~al.(2016)He, Huang, Qiao, Loy, and Tang]{he2016reading}
P.~He, W.~Huang, Y.~Qiao, C.~C. Loy, and X.~Tang.
\newblock Reading scene text in deep convolutional sequences.
\newblock In \emph{Thirtieth AAAI conference on artificial intelligence}, 2016.

\bibitem[He et~al.(2017{\natexlab{c}})He, Huang, He, Zhu, Qiao, and
  Li]{He2017SSTD}
P.~He, W.~Huang, T.~He, Q.~Zhu, Y.~Qiao, and X.~Li.
\newblock Single shot text detector with regional attention.
\newblock In \emph{The IEEE International Conference on Computer Vision
  (ICCV)}, 2017{\natexlab{c}}.

\bibitem[He et~al.(2018)He, Tian, Huang, Shen, Qiao, and Sun]{he2018end}
T.~He, Z.~Tian, W.~Huang, C.~Shen, Y.~Qiao, and C.~Sun.
\newblock An end-to-end textspotter with explicit alignment and attention.
\newblock In \emph{Proceedings of the IEEE Conference on Computer Vision and
  Pattern Recognition (CVPR)}, pages 5020--5029, 2018.

\bibitem[He et~al.(2017{\natexlab{d}})He, Zhang, Yin, and Liu]{He_2017_ICCV}
W.~He, X.-Y. Zhang, F.~Yin, and C.-L. Liu.
\newblock Deep direct regression for multi-oriented scene text detection.
\newblock In \emph{The IEEE International Conference on Computer Vision
  (ICCV)}, 2017{\natexlab{d}}.

\bibitem[He et~al.(2005)He, Liu, Ma, and Li]{he2005new}
Z.~He, J.~Liu, H.~Ma, and P.~Li.
\newblock A new automatic extraction method of container identity codes.
\newblock \emph{IEEE Transactions on intelligent transportation systems},
  6\penalty0 (1):\penalty0 72--78, 2005.

\bibitem[Hochreiter and Schmidhuber(1997)]{hochreiter1997long}
S.~Hochreiter and J.~Schmidhuber.
\newblock Long short-term memory.
\newblock \emph{Neural computation}, 9\penalty0 (8):\penalty0 1735--1780, 1997.

\bibitem[Hu et~al.(2017)Hu, Zhang, Luo, Wang, Han, and Ding]{Hu2018wordsup}
H.~Hu, C.~Zhang, Y.~Luo, Y.~Wang, J.~Han, and E.~Ding.
\newblock Wordsup: Exploiting word annotations for character based text
  detection.
\newblock In \emph{Proceedings of the IEEE International Conference on Computer
  Vision. 2017.}, 2017.

\bibitem[Huang et~al.(2013)Huang, Lin, Yang, and Wang]{huang2013text}
W.~Huang, Z.~Lin, J.~Yang, and J.~Wang.
\newblock Text localization in natural images using stroke feature transform
  and text covariance descriptors.
\newblock In \emph{Proceedings of the IEEE International Conference on Computer
  Vision}, pages 1241--1248, 2013.

\bibitem[Huang et~al.(2014)Huang, Qiao, and Tang]{huang2014robust}
W.~Huang, Y.~Qiao, and X.~Tang.
\newblock Robust scene text detection with convolution neural network induced
  mser trees.
\newblock In \emph{European conference on computer vision}, pages 497--511.
  Springer, 2014.

\bibitem[Jaderberg et~al.(2014{\natexlab{a}})Jaderberg, Simonyan, Vedaldi, and
  Zisserman]{jaderberg2014deep}
M.~Jaderberg, K.~Simonyan, A.~Vedaldi, and A.~Zisserman.
\newblock Deep structured output learning for unconstrained text recognition.
\newblock \emph{ICLR2015}, 2014{\natexlab{a}}.

\bibitem[Jaderberg et~al.(2014{\natexlab{b}})Jaderberg, Simonyan, Vedaldi, and
  Zisserman]{jaderberg2014synthetic}
M.~Jaderberg, K.~Simonyan, A.~Vedaldi, and A.~Zisserman.
\newblock Synthetic data and artificial neural networks for natural scene text
  recognition.
\newblock \emph{arXiv preprint arXiv:1406.2227}, 2014{\natexlab{b}}.

\bibitem[Jaderberg et~al.(2014{\natexlab{c}})Jaderberg, Vedaldi, and
  Zisserman]{jaderberg2014deepf}
M.~Jaderberg, A.~Vedaldi, and A.~Zisserman.
\newblock Deep features for text spotting.
\newblock In \emph{In Proceedings of European Conference on Computer Vision
  (ECCV)}, pages 512--528. Springer, 2014{\natexlab{c}}.

\bibitem[Jaderberg et~al.(2015)Jaderberg, Simonyan, Zisserman,
  et~al.]{jaderberg2015spatial}
M.~Jaderberg, K.~Simonyan, A.~Zisserman, et~al.
\newblock Spatial transformer networks.
\newblock In \emph{Advances in neural information processing systems}, pages
  2017--2025, 2015.

\bibitem[Jaderberg et~al.(2016)Jaderberg, Simonyan, Vedaldi, and
  Zisserman]{jaderberg2016reading}
M.~Jaderberg, K.~Simonyan, A.~Vedaldi, and A.~Zisserman.
\newblock Reading text in the wild with convolutional neural networks.
\newblock \emph{International Journal of Computer Vision}, 116\penalty0
  (1):\penalty0 1--20, 2016.

\bibitem[Jain and Yu(1998)]{jain1998automatic}
A.~K. Jain and B.~Yu.
\newblock Automatic text location in images and video frames.
\newblock \emph{Pattern recognition}, 31\penalty0 (12):\penalty0 2055--2076,
  1998.

\bibitem[Jiang et~al.(2017)Jiang, Zhu, Wang, Yang, Li, Wang, Fu, and
  Luo]{jiang2017r2cnn}
Y.~Jiang, X.~Zhu, X.~Wang, S.~Yang, W.~Li, H.~Wang, P.~Fu, and Z.~Luo.
\newblock R2cnn: rotational region cnn for orientation robust scene text
  detection.
\newblock \emph{arXiv preprint arXiv:1706.09579}, 2017.

\bibitem[Jung et~al.(2004)Jung, Kim, and Jain]{jung2004text}
K.~Jung, K.~I. Kim, and A.~K. Jain.
\newblock Text information extraction in images and video: a survey.
\newblock \emph{Pattern recognition}, 37\penalty0 (5):\penalty0 977--997, 2004.

\bibitem[Kang et~al.(2014)Kang, Li, and Doermann]{kang2014orientation}
L.~Kang, Y.~Li, and D.~Doermann.
\newblock Orientation robust text line detection in natural images.
\newblock In \emph{Proceedings of the IEEE Conference on Computer Vision and
  Pattern Recognition (CVPR)}, pages 4034--4041, 2014.

\bibitem[Karatzas and Antonacopoulos(2004)]{karatzas2004text}
D.~Karatzas and A.~Antonacopoulos.
\newblock Text extraction from web images based on a split-and-merge
  segmentation method using colour perception.
\newblock In \emph{Proceedings of the 17th International Conference on Pattern
  Recognition, 2004. ICPR 2004.}, volume~2, pages 634--637. IEEE, 2004.

\bibitem[Karatzas et~al.(2013)Karatzas, Shafait, Uchida, Iwamura, i~Bigorda,
  Mestre, Mas, Mota, Almazan, and de~las Heras]{karatzas2013icdar}
D.~Karatzas, F.~Shafait, S.~Uchida, M.~Iwamura, L.~G. i~Bigorda, S.~R. Mestre,
  J.~Mas, D.~F. Mota, J.~A. Almazan, and L.~P. de~las Heras.
\newblock Icdar 2013 robust reading competition.
\newblock In \emph{2013 12th International Conference on Document Analysis and
  Recognition (ICDAR)}, pages 1484--1493. IEEE, 2013.

\bibitem[Karatzas et~al.(2015)Karatzas, Gomez-Bigorda, Nicolaou, Ghosh,
  Bagdanov, Iwamura, Matas, Neumann, Chandrasekhar, Lu,
  et~al.]{karatzas2015icdar}
D.~Karatzas, L.~Gomez-Bigorda, A.~Nicolaou, S.~Ghosh, A.~Bagdanov, M.~Iwamura,
  J.~Matas, L.~Neumann, V.~R. Chandrasekhar, S.~Lu, et~al.
\newblock Icdar 2015 competition on robust reading.
\newblock In \emph{2015 13th International Conference on Document Analysis and
  Recognition (ICDAR)}, pages 1156--1160. IEEE, 2015.

\bibitem[Kipf and Welling(2016)]{kipf2016semi}
T.~N. Kipf and M.~Welling.
\newblock Semi-supervised classification with graph convolutional networks.
\newblock \emph{arXiv preprint arXiv:1609.02907}, 2016.

\bibitem[Krizhevsky et~al.(2012)Krizhevsky, Sutskever, and
  Hinton]{krizhevsky2012imagenet}
A.~Krizhevsky, I.~Sutskever, and G.~E. Hinton.
\newblock Imagenet classification with deep convolutional neural networks.
\newblock In \emph{Advances in neural information processing systems}, pages
  1097--1105, 2012.

\bibitem[Lee and Osindero(2016)]{lee2016recursive}
C.-Y. Lee and S.~Osindero.
\newblock Recursive recurrent nets with attention modeling for ocr in the wild.
\newblock In \emph{Proceedings of the IEEE Conference on Computer Vision and
  Pattern Recognition (CVPR)}, pages 2231--2239, 2016.

\bibitem[Lee et~al.(2011)Lee, Lee, Lee, Yuille, and Koch]{lee2011adaboost}
J.-J. Lee, P.-H. Lee, S.-W. Lee, A.~Yuille, and C.~Koch.
\newblock Adaboost for text detection in natural scene.
\newblock In \emph{2011 International Conference on Document Analysis and
  Recognition (ICDAR)}, pages 429--434. IEEE, 2011.

\bibitem[Lee and Kim(2013)]{lee2013integrating}
S.~Lee and J.~H. Kim.
\newblock Integrating multiple character proposals for robust scene text
  extraction.
\newblock \emph{Image and Vision Computing}, 31\penalty0 (11):\penalty0
  823--840, 2013.

\bibitem[Li et~al.(2017{\natexlab{a}})Li, Wang, and Shen]{Li_2017_ICCV}
H.~Li, P.~Wang, and C.~Shen.
\newblock Towards end-to-end text spotting with convolutional recurrent neural
  networks.
\newblock In \emph{The IEEE International Conference on Computer Vision
  (ICCV)}, 2017{\natexlab{a}}.

\bibitem[Li et~al.(2019)Li, Wang, Shen, and Zhang]{li2018show}
H.~Li, P.~Wang, C.~Shen, and G.~Zhang.
\newblock Show, attend and read: A simple and strong baseline for irregular
  text recognition.
\newblock \emph{AAAI}, 2019.

\bibitem[Li et~al.(2017{\natexlab{b}})Li, En, Li, and Zhang]{rong2017weakly}
R.~Li, M.~En, J.~Li, and H.~Zhang.
\newblock weakly supervised text attention network for generating text
  proposals in scene images.
\newblock In \emph{2017 14th IAPR International Conference on Document Analysis
  and Recognition (ICDAR)}, volume~1, pages 324--330. IEEE, 2017{\natexlab{b}}.

\bibitem[Liao et~al.(2017)Liao, Shi, Bai, Wang, and Liu]{liao2017textboxes}
M.~Liao, B.~Shi, X.~Bai, X.~Wang, and W.~Liu.
\newblock Textboxes: A fast text detector with a single deep neural network.
\newblock In \emph{AAAI}, pages 4161--4167, 2017.

\bibitem[Liao et~al.(2018{\natexlab{a}})Liao, Shi, and
  Bai]{liao2018textboxes++}
M.~Liao, B.~Shi, and X.~Bai.
\newblock Textboxes++: A single-shot oriented scene text detector.
\newblock \emph{IEEE Transactions on Image Processing}, 27\penalty0
  (8):\penalty0 3676--3690, 2018{\natexlab{a}}.

\bibitem[Liao et~al.(2018{\natexlab{b}})Liao, Zhu, Shi, Xia, and
  Bai]{liao2018rotation}
M.~Liao, Z.~Zhu, B.~Shi, G.-s. Xia, and X.~Bai.
\newblock Rotation-sensitive regression for oriented scene text detection.
\newblock In \emph{Proceedings of the IEEE Conference on Computer Vision and
  Pattern Recognition (CVPR)}, pages 5909--5918, 2018{\natexlab{b}}.

\bibitem[Liao et~al.(2019{\natexlab{a}})Liao, Song, He, Long, Yao, and
  Bai]{liao2019synthtext3d}
M.~Liao, B.~Song, M.~He, S.~Long, C.~Yao, and X.~Bai.
\newblock Synthtext3d: Synthesizing scene text images from 3d virtual worlds.
\newblock \emph{arXiv preprint arXiv:1907.06007}, 2019{\natexlab{a}}.

\bibitem[Liao et~al.(2019{\natexlab{b}})Liao, Zhang, Wan, Xie, Liang, Lyu, Yao,
  and Bai]{liao2018scene}
M.~Liao, J.~Zhang, Z.~Wan, F.~Xie, J.~Liang, P.~Lyu, C.~Yao, and X.~Bai.
\newblock Scene text recognition from two-dimensional perspective.
\newblock \emph{AAAI}, 2019{\natexlab{b}}.

\bibitem[Liu et~al.(2015)Liu, Shen, and Lin]{liu2015deep}
F.~Liu, C.~Shen, and G.~Lin.
\newblock Deep convolutional neural fields for depth estimation from a single
  image.
\newblock In \emph{Proceedings of the IEEE Conference on Computer Vision and
  Pattern Recognition (CVPR)}, pages 5162--5170, 2015.

\bibitem[Liu et~al.(2018{\natexlab{a}})Liu, Ouyang, Wang, Fieguth, Chen, Liu,
  and Pietik{\"a}inen]{liu2018deep}
L.~Liu, W.~Ouyang, X.~Wang, P.~Fieguth, J.~Chen, X.~Liu, and
  M.~Pietik{\"a}inen.
\newblock Deep learning for generic object detection: A survey.
\newblock \emph{arXiv preprint arXiv:1809.02165}, 2018{\natexlab{a}}.

\bibitem[Liu et~al.(2016{\natexlab{a}})Liu, Anguelov, Erhan, Szegedy, Reed, Fu,
  and Berg]{liu2016ssd}
W.~Liu, D.~Anguelov, D.~Erhan, C.~Szegedy, S.~Reed, C.-Y. Fu, and A.~C. Berg.
\newblock {SSD}: Single shot multibox detector.
\newblock In \emph{In Proceedings of European Conference on Computer Vision
  (ECCV)}, pages 21--37. Springer, 2016{\natexlab{a}}.

\bibitem[Liu et~al.(2016{\natexlab{b}})Liu, Chen, Wong, Su, and
  Han]{liu2016star}
W.~Liu, C.~Chen, K.-Y.~K. Wong, Z.~Su, and J.~Han.
\newblock Star-net: A spatial attention residue network for scene text
  recognition.
\newblock In \emph{BMVC}, volume~2, page~7, 2016{\natexlab{b}}.

\bibitem[Liu et~al.(2018{\natexlab{b}})Liu, Chen, and Wong]{liu2018char}
W.~Liu, C.~Chen, and K.~Wong.
\newblock Char-net: A character-aware neural network for distorted scene text
  recognition.
\newblock In \emph{AAAI Conference on Artificial Intelligence}. New Orleans,
  Louisiana, USA, 2018{\natexlab{b}}.

\bibitem[Liu(1975)]{liu1975old}
X.~Liu.
\newblock Old book of tang.
\newblock \emph{Beijing: Zhonghua Book Company}, 1975.

\bibitem[Liu and Samarabandu(2005{\natexlab{a}})]{liu2005edge}
X.~Liu and J.~Samarabandu.
\newblock An edge-based text region extraction algorithm for indoor mobile
  robot navigation.
\newblock In \emph{Mechatronics and Automation, 2005 IEEE International
  Conference}, volume~2, pages 701--706. IEEE, 2005{\natexlab{a}}.

\bibitem[Liu and Samarabandu(2005{\natexlab{b}})]{liu2005simple}
X.~Liu and J.~K. Samarabandu.
\newblock A simple and fast text localization algorithm for indoor mobile robot
  navigation.
\newblock In \emph{Image Processing: Algorithms and Systems IV}, volume 5672,
  pages 139--151. International Society for Optics and Photonics,
  2005{\natexlab{b}}.

\bibitem[Liu et~al.(2018{\natexlab{c}})Liu, Liang, Yan, Chen, Qiao, and
  Yan]{liu2018fots}
X.~Liu, D.~Liang, S.~Yan, D.~Chen, Y.~Qiao, and J.~Yan.
\newblock Fots: Fast oriented text spotting with a unified network.
\newblock \emph{CVPR2018}, 2018{\natexlab{c}}.

\bibitem[Liu and Jin(2017)]{Liu2017Deep}
Y.~Liu and L.~Jin.
\newblock Deep matching prior network: Toward tighter multi-oriented text
  detection.
\newblock 2017.

\bibitem[Liu et~al.(2017)Liu, Jin, Zhang, and Zhang]{Yuliang2017Detecting}
Y.~Liu, L.~Jin, S.~Zhang, and S.~Zhang.
\newblock Detecting curve text in the wild: New dataset and new solution.
\newblock \emph{arXiv preprint arXiv:1712.02170}, 2017.

\bibitem[Liu et~al.(2019)Liu, Jin, Xie, Luo, Zhang, and Xie]{liu2019tightness}
Y.~Liu, L.~Jin, Z.~Xie, C.~Luo, S.~Zhang, and L.~Xie.
\newblock Tightness-aware evaluation protocol for scene text detection.
\newblock In \emph{Proceedings of the IEEE Conference on Computer Vision and
  Pattern Recognition}, pages 9612--9620, 2019.

\bibitem[Liu et~al.(2018{\natexlab{d}})Liu, Li, Ren, Yu, and
  Goh]{liu2018squeezedtext}
Z.~Liu, Y.~Li, F.~Ren, H.~Yu, and W.~Goh.
\newblock Squeezedtext: A real-time scene text recognition by binary
  convolutional encoder-decoder network.
\newblock \emph{AAAI}, 2018{\natexlab{d}}.

\bibitem[Liu et~al.(2018{\natexlab{e}})Liu, Lin, Yang, Feng, Lin, and
  Ling~Goh]{liu2018learning}
Z.~Liu, G.~Lin, S.~Yang, J.~Feng, W.~Lin, and W.~Ling~Goh.
\newblock Learning markov clustering networks for scene text detection.
\newblock In \emph{Proceedings of the IEEE Conference on Computer Vision and
  Pattern Recognition (CVPR)}, pages 6936--6944, 2018{\natexlab{e}}.

\bibitem[Long and Yao(2020)]{long2020unrealtext}
S.~Long and C.~Yao.
\newblock Unrealtext: Synthesizing realistic scene text images from the unreal
  world.
\newblock \emph{arXiv preprint arXiv:2003.10608}, 2020.

\bibitem[Long et~al.(2018)Long, Ruan, Zhang, He, Wu, and
  Yao]{long2018textsnake}
S.~Long, J.~Ruan, W.~Zhang, X.~He, W.~Wu, and C.~Yao.
\newblock Textsnake: A flexible representation for detecting text of arbitrary
  shapes.
\newblock In \emph{In Proceedings of European Conference on Computer Vision
  (ECCV)}, 2018.

\bibitem[Long et~al.(2019)Long, Guan, Wang, Bian, and Yao]{long2019alchemy}
S.~Long, Y.~Guan, B.~Wang, K.~Bian, and C.~Yao.
\newblock Alchemy: Techniques for rectification based irregular scene text
  recognition.
\newblock \emph{arXiv preprint arXiv:1908.11834}, 2019.

\bibitem[Long et~al.(2020)Long, Guan, Bian, and Yao]{long2020new}
S.~Long, Y.~Guan, K.~Bian, and C.~Yao.
\newblock A new perspective for flexible feature gathering in scene text
  recognition via character anchor pooling.
\newblock In \emph{ICASSP 2020-2020 IEEE International Conference on Acoustics,
  Speech and Signal Processing (ICASSP)}, pages 2458--2462. IEEE, 2020.

\bibitem[Lyu et~al.(2018{\natexlab{a}})Lyu, Liao, Yao, Wu, and
  Bai]{lyu2018mask}
P.~Lyu, M.~Liao, C.~Yao, W.~Wu, and X.~Bai.
\newblock Mask textspotter: An end-to-end trainable neural network for spotting
  text with arbitrary shapes.
\newblock In \emph{In Proceedings of European Conference on Computer Vision
  (ECCV)}, 2018{\natexlab{a}}.

\bibitem[Lyu et~al.(2018{\natexlab{b}})Lyu, Yao, Wu, Yan, and Bai]{Lyu2018}
P.~Lyu, C.~Yao, W.~Wu, S.~Yan, and X.~Bai.
\newblock Multi-oriented scene text detection via corner localization and
  region segmentation.
\newblock In \emph{2018 IEEE Conference on Computer Vision and Pattern
  Recognition (CVPR)}, 2018{\natexlab{b}}.

\bibitem[Ma et~al.(2017)Ma, Shao, Ye, Wang, Wang, Zheng, and
  Xue]{ma2017arbitrary}
J.~Ma, W.~Shao, H.~Ye, L.~Wang, H.~Wang, Y.~Zheng, and X.~Xue.
\newblock Arbitrary-oriented scene text detection via rotation proposals.
\newblock In \emph{IEEE Transactions on Multimedia, 2018}, 2017.

\bibitem[Mammeri et~al.(2014)Mammeri, Khiari, and Boukerche]{mammeri2014road}
A.~Mammeri, E.-H. Khiari, and A.~Boukerche.
\newblock Road-sign text recognition architecture for intelligent
  transportation systems.
\newblock In \emph{2014 IEEE 80th Vehicular Technology Conference (VTC Fall)},
  pages 1--5. IEEE, 2014.

\bibitem[Mammeri et~al.(2016)Mammeri, Boukerche, et~al.]{mammeri2016mser}
A.~Mammeri, A.~Boukerche, et~al.
\newblock Mser-based text detection and communication algorithm for autonomous
  vehicles.
\newblock In \emph{2016 IEEE Symposium on Computers and Communication (ISCC)},
  pages 1218--1223. IEEE, 2016.

\bibitem[Mishra et~al.(2011)Mishra, Alahari, and Jawahar]{mishra2011mrf}
A.~Mishra, K.~Alahari, and C.~Jawahar.
\newblock An mrf model for binarization of natural scene text.
\newblock In \emph{ICDAR-International Conference on Document Analysis and
  Recognition}. IEEE, 2011.

\bibitem[Mishra et~al.(2012)Mishra, Alahari, and Jawahar]{mishra2012scene}
A.~Mishra, K.~Alahari, and C.~Jawahar.
\newblock Scene text recognition using higher order language priors.
\newblock In \emph{BMVC-British Machine Vision Conference}. BMVA, 2012.

\bibitem[Neumann and Matas(2010)]{neumann2010method}
L.~Neumann and J.~Matas.
\newblock A method for text localization and recognition in real-world images.
\newblock In \emph{Asian Conference on Computer Vision}, pages 770--783.
  Springer, 2010.

\bibitem[Neumann and Matas(2012)]{neumann2012real}
L.~Neumann and J.~Matas.
\newblock Real-time scene text localization and recognition.
\newblock In \emph{2012 IEEE Conference on Computer Vision and Pattern
  Recognition (CVPR)}, pages 3538--3545. IEEE, 2012.

\bibitem[Neumann and Matas(2013)]{neumann2013combining}
L.~Neumann and J.~Matas.
\newblock On combining multiple segmentations in scene text recognition.
\newblock In \emph{2013 12th International Conference on Document Analysis and
  Recognition (ICDAR)}, pages 523--527. IEEE, 2013.

\bibitem[Nomura et~al.(2005)Nomura, Yamanaka, Katai, Kawakami, and
  Shiose]{nomura2005novel}
S.~Nomura, K.~Yamanaka, O.~Katai, H.~Kawakami, and T.~Shiose.
\newblock A novel adaptive morphological approach for degraded character image
  segmentation.
\newblock \emph{Pattern Recognition}, 38\penalty0 (11):\penalty0 1961--1975,
  2005.

\bibitem[Parkinson et~al.(2016)Parkinson, Jacobsen, Ferguson, and
  Pombo]{parkinson2016instant}
C.~Parkinson, J.~J. Jacobsen, D.~B. Ferguson, and S.~A. Pombo.
\newblock Instant translation system, Nov.~29 2016.
\newblock US Patent 9,507,772.

\bibitem[Qin et~al.(2019)Qin, Bissacco, Raptis, Fujii, and
  Xiao]{qin2019towards}
S.~Qin, A.~Bissacco, M.~Raptis, Y.~Fujii, and Y.~Xiao.
\newblock Towards unconstrained end-to-end text spotting.
\newblock In \emph{Proceedings of the IEEE International Conference on Computer
  Vision}, pages 4704--4714, 2019.

\bibitem[Qiu et~al.(2017)Qiu, Zhong, Zhang, Qiao, Xiao, Kim, and
  Wang]{qiu2017unrealcv}
W.~Qiu, F.~Zhong, Y.~Zhang, S.~Qiao, Z.~Xiao, T.~S. Kim, and Y.~Wang.
\newblock Unrealcv: Virtual worlds for computer vision.
\newblock In \emph{Proceedings of the 25th ACM international conference on
  Multimedia}, pages 1221--1224. ACM, 2017.

\bibitem[Quy~Phan et~al.(2013)Quy~Phan, Shivakumara, Tian, and
  Lim~Tan]{quy2013recognizing}
T.~Quy~Phan, P.~Shivakumara, S.~Tian, and C.~Lim~Tan.
\newblock Recognizing text with perspective distortion in natural scenes.
\newblock In \emph{Proceedings of the IEEE International Conference on Computer
  Vision (ICCV)}, pages 569--576, 2013.

\bibitem[Redmon and Farhadi(2017)]{redmon2017yolo9000}
J.~Redmon and A.~Farhadi.
\newblock Yolo9000: better, faster, stronger.
\newblock \emph{arXiv preprint}, 2017.

\bibitem[Redmon et~al.(2016)Redmon, Divvala, Girshick, and
  Farhadi]{redmon2016you}
J.~Redmon, S.~Divvala, R.~Girshick, and A.~Farhadi.
\newblock You only look once: Unified, real-time object detection.
\newblock In \emph{Proceedings of the IEEE Conference on Computer Vision and
  Pattern Recognition (CVPR)}, pages 779--788, 2016.

\bibitem[Ren et~al.(2015)Ren, He, Girshick, and Sun]{ren2015faster}
S.~Ren, K.~He, R.~Girshick, and J.~Sun.
\newblock Faster r-cnn: Towards real-time object detection with region proposal
  networks.
\newblock In \emph{Advances in neural information processing systems}, pages
  91--99, 2015.

\bibitem[Rodriguez-Serrano et~al.(2013)Rodriguez-Serrano, Perronnin, and
  Meylan]{rodriguez2013label}
J.~A. Rodriguez-Serrano, F.~Perronnin, and F.~Meylan.
\newblock Label embedding for text recognition.
\newblock In \emph{Proceedings of the British Machine Vision Conference}.
  Citeseer, 2013.

\bibitem[Rodriguez-Serrano et~al.(2015)Rodriguez-Serrano, Gordo, and
  Perronnin]{rodriguez2015label}
J.~A. Rodriguez-Serrano, A.~Gordo, and F.~Perronnin.
\newblock Label embedding: A frugal baseline for text recognition.
\newblock \emph{International Journal of Computer Vision}, 113\penalty0
  (3):\penalty0 193--207, 2015.

\bibitem[Ronneberger et~al.(2015)Ronneberger, Fischer, and
  Brox]{Ronneberger2015U}
O.~Ronneberger, P.~Fischer, and T.~Brox.
\newblock \emph{U-Net: Convolutional Networks for Biomedical Image
  Segmentation}.
\newblock Springer International Publishing, 2015.

\bibitem[Roy et~al.(2009)Roy, Pal, Llados, and Delalandre]{roy2009multi}
P.~P. Roy, U.~Pal, J.~Llados, and M.~Delalandre.
\newblock Multi-oriented and multi-sized touching character segmentation using
  dynamic programming.
\newblock In \emph{10th International Conference onDocument Analysis and
  Recognition, 2009}. IEEE, 2009.

\bibitem[Russakovsky et~al.(2015)Russakovsky, Deng, Su, Krause, Satheesh, Ma,
  Huang, Karpathy, Khosla, Bernstein, et~al.]{russakovsky2015imagenet}
O.~Russakovsky, J.~Deng, H.~Su, J.~Krause, S.~Satheesh, S.~Ma, Z.~Huang,
  A.~Karpathy, A.~Khosla, M.~Bernstein, et~al.
\newblock Imagenet large scale visual recognition challenge.
\newblock \emph{International Journal of Computer Vision}, 115\penalty0
  (3):\penalty0 211--252, 2015.

\bibitem[Schroth et~al.(2011)Schroth, Hilsenbeck, Huitl, Schweiger, and
  Steinbach]{schroth2011exploiting}
G.~Schroth, S.~Hilsenbeck, R.~Huitl, F.~Schweiger, and E.~Steinbach.
\newblock Exploiting text-related features for content-based image retrieval.
\newblock In \emph{2011 IEEE International Symposium on Multimedia}, pages
  77--84. IEEE, 2011.

\bibitem[Schulz et~al.(2015)Schulz, Talbot, Lam, Dayoub, Corke, Upcroft, and
  Wyeth]{schulz2015robot}
R.~Schulz, B.~Talbot, O.~Lam, F.~Dayoub, P.~Corke, B.~Upcroft, and G.~Wyeth.
\newblock Robot navigation using human cues: A robot navigation system for
  symbolic goal-directed exploration.
\newblock In \emph{Proceedings of the 2015 IEEE International Conference on
  Robotics and Automation (ICRA 2015)}, pages 1100--1105. IEEE, 2015.

\bibitem[Sheshadri and Divvala(2012)]{sheshadri2012exemplar}
K.~Sheshadri and S.~K. Divvala.
\newblock Exemplar driven character recognition in the wild.
\newblock In \emph{BMVC}, pages 1--10, 2012.

\bibitem[Shi et~al.(2016)Shi, Wang, Lyu, Yao, and Bai]{shi2016robust}
B.~Shi, X.~Wang, P.~Lyu, C.~Yao, and X.~Bai.
\newblock Robust scene text recognition with automatic rectification.
\newblock In \emph{Proceedings of the IEEE Conference on Computer Vision and
  Pattern Recognition (CVPR)}, pages 4168--4176, 2016.

\bibitem[Shi et~al.(2017{\natexlab{a}})Shi, Bai, and Belongie]{Shi_2017_CVPR}
B.~Shi, X.~Bai, and S.~Belongie.
\newblock Detecting oriented text in natural images by linking segments.
\newblock In \emph{The IEEE Conference on Computer Vision and Pattern
  Recognition (CVPR)}, 2017{\natexlab{a}}.

\bibitem[Shi et~al.(2017{\natexlab{b}})Shi, Bai, and Yao]{shi2017end}
B.~Shi, X.~Bai, and C.~Yao.
\newblock An end-to-end trainable neural network for image-based sequence
  recognition and its application to scene text recognition.
\newblock \emph{IEEE transactions on pattern analysis and machine
  intelligence}, 39\penalty0 (11):\penalty0 2298--2304, 2017{\natexlab{b}}.

\bibitem[Shi et~al.(2018)Shi, Yang, Wang, Lyu, Bai, and Yao]{shi2018aster}
B.~Shi, M.~Yang, X.~Wang, P.~Lyu, X.~Bai, and C.~Yao.
\newblock Aster: An attentional scene text recognizer with flexible
  rectification.
\newblock \emph{IEEE transactions on pattern analysis and machine
  intelligence}, 31\penalty0 (11):\penalty0 855--868, 2018.

\bibitem[Shi et~al.(2013)Shi, Wang, Xiao, Zhang, Gao, and Zhang]{shi2013scene}
C.~Shi, C.~Wang, B.~Xiao, Y.~Zhang, S.~Gao, and Z.~Zhang.
\newblock Scene text recognition using part-based tree-structured character
  detection.
\newblock In \emph{2013 IEEE Conference on Computer Vision and Pattern
  Recognition (CVPR)}, pages 2961--2968. IEEE, 2013.

\bibitem[Shivakumara et~al.(2011)Shivakumara, Bhowmick, Su, Tan, and
  Pal]{shivakumara2011new}
P.~Shivakumara, S.~Bhowmick, B.~Su, C.~L. Tan, and U.~Pal.
\newblock A new gradient based character segmentation method for video text
  recognition.
\newblock In \emph{2011 International Conference on Document Analysis and
  Recognition (ICDAR)}. IEEE, 2011.

\bibitem[Su and Lu(2014)]{su2014accurate}
B.~Su and S.~Lu.
\newblock Accurate scene text recognition based on recurrent neural network.
\newblock In \emph{Asian Conference on Computer Vision}, pages 35--48.
  Springer, 2014.

\bibitem[Sun et~al.(2019)Sun, Liu, Liu, Han, Ding, and Liu]{sun2019chinese}
Y.~Sun, J.~Liu, W.~Liu, J.~Han, E.~Ding, and J.~Liu.
\newblock Chinese street view text: Large-scale chinese text reading with
  partially supervised learning.
\newblock In \emph{Proceedings of the IEEE International Conference on Computer
  Vision}, pages 9086--9095, 2019.

\bibitem[Sutskever et~al.(2014)Sutskever, Vinyals, and
  Le]{sutskever2014sequence}
I.~Sutskever, O.~Vinyals, and Q.~V. Le.
\newblock Sequence to sequence learning with neural networks.
\newblock In \emph{Advances in neural information processing systems}, pages
  3104--3112, 2014.

\bibitem[Tian et~al.(2015)Tian, Pan, Huang, Lu, Yu, and Lim~Tan]{tian2015text}
S.~Tian, Y.~Pan, C.~Huang, S.~Lu, K.~Yu, and C.~Lim~Tan.
\newblock Text flow: A unified text detection system in natural scene images.
\newblock In \emph{Proceedings of the IEEE international conference on computer
  vision}, pages 4651--4659, 2015.

\bibitem[Tian et~al.(2017)Tian, Lu, and Li]{tian2017wetext}
S.~Tian, S.~Lu, and C.~Li.
\newblock Wetext: Scene text detection under weak supervision.
\newblock In \emph{Proc. ICCV}, 2017.

\bibitem[Tian et~al.(2016)Tian, Huang, He, He, and Qiao]{tian2016detecting}
Z.~Tian, W.~Huang, T.~He, P.~He, and Y.~Qiao.
\newblock Detecting text in natural image with connectionist text proposal
  network.
\newblock In \emph{In Proceedings of European Conference on Computer Vision
  (ECCV)}, pages 56--72. Springer, 2016.

\bibitem[Tian et~al.(2019)Tian, Shu, Lyu, Li, Zhou, Shen, and
  Jia]{tian2019learning}
Z.~Tian, M.~Shu, P.~Lyu, R.~Li, C.~Zhou, X.~Shen, and J.~Jia.
\newblock Learning shape-aware embedding for scene text detection.
\newblock In \emph{Proceedings of the IEEE Conference on Computer Vision and
  Pattern Recognition}, pages 4234--4243, 2019.

\bibitem[Tsai et~al.(2011)Tsai, Chen, Chen, Schroth, Grzeszczuk, and
  Girod]{tsai2011mobile}
S.~S. Tsai, H.~Chen, D.~Chen, G.~Schroth, R.~Grzeszczuk, and B.~Girod.
\newblock Mobile visual search on printed documents using text and low bit-rate
  features.
\newblock In \emph{18th IEEE International Conference on Image Processing
  (ICIP)}, pages 2601--2604. IEEE, 2011.

\bibitem[Tu et~al.(2012)Tu, Ma, Liu, Bai, and Yao]{tu2012detecting}
Z.~Tu, Y.~Ma, W.~Liu, X.~Bai, and C.~Yao.
\newblock Detecting texts of arbitrary orientations in natural images.
\newblock In \emph{2012 IEEE Conference on Computer Vision and Pattern
  Recognition}, pages 1083--1090. IEEE, 2012.

\bibitem[Uchida(2014)]{uchida2014text}
S.~Uchida.
\newblock Text localization and recognition in images and video.
\newblock In \emph{Handbook of Document Image Processing and Recognition},
  pages 843--883. Springer, 2014.

\bibitem[Wachenfeld et~al.(2006)Wachenfeld, Klein, and
  Jiang]{wachenfeld2006recognition}
S.~Wachenfeld, H.-U. Klein, and X.~Jiang.
\newblock Recognition of screen-rendered text.
\newblock In \emph{18th International Conference on Pattern Recognition, 2006.
  ICPR 2006.}, volume~2, pages 1086--1089. IEEE, 2006.

\bibitem[Wakahara and Kita(2011)]{wakahara2011binarization}
T.~Wakahara and K.~Kita.
\newblock Binarization of color character strings in scene images using k-means
  clustering and support vector machines.
\newblock In \emph{2011 International Conference on Document Analysis and
  Recognition (ICDAR)}, pages 274--278. IEEE, 2011.

\bibitem[Wang et~al.(2017)Wang, Yin, and Liu]{wang2017scene}
C.~Wang, F.~Yin, and C.-L. Liu.
\newblock Scene text detection with novel superpixel based character candidate
  extraction.
\newblock In \emph{2017 14th IAPR International Conference on Document Analysis
  and Recognition (ICDAR)}, volume~1, pages 929--934. IEEE, 2017.

\bibitem[Wang et~al.(2018)Wang, Zhao, Li, Wang, and Tao]{wang2018geometry}
F.~Wang, L.~Zhao, X.~Li, X.~Wang, and D.~Tao.
\newblock Geometry-aware scene text detection with instance transformation
  network.
\newblock In \emph{Proceedings of the IEEE Conference on Computer Vision and
  Pattern Recognition (CVPR)}, pages 1381--1389, 2018.

\bibitem[Wang et~al.(2011)Wang, Babenko, and Belongie]{wang2011end}
K.~Wang, B.~Babenko, and S.~Belongie.
\newblock End-to-end scene text recognition.
\newblock In \emph{2011 IEEE International Conference on Computer Vision
  (ICCV),}, pages 1457--1464. IEEE, 2011.

\bibitem[Wang et~al.(2012)Wang, Wu, Coates, and Ng]{wang2012end}
T.~Wang, D.~J. Wu, A.~Coates, and A.~Y. Ng.
\newblock End-to-end text recognition with convolutional neural networks.
\newblock In \emph{2012 21st International Conference on Pattern Recognition
  (ICPR)}, pages 3304--3308. IEEE, 2012.

\bibitem[Wang et~al.(2019{\natexlab{a}})Wang, Xie, Li, Hou, Lu, Yu, and
  Shao]{wang2019shape}
W.~Wang, E.~Xie, X.~Li, W.~Hou, T.~Lu, G.~Yu, and S.~Shao.
\newblock Shape robust text detection with progressive scale expansion network.
\newblock \emph{Proceedings of the IEEE Conference on Computer Vision and
  Pattern Recognition (CVPR)}, 2019{\natexlab{a}}.

\bibitem[Wang et~al.(2019{\natexlab{b}})Wang, Jiang, Luo, Liu, Choi, and
  Kim]{wang2019arbitrary}
X.~Wang, Y.~Jiang, Z.~Luo, C.-L. Liu, H.~Choi, and S.~Kim.
\newblock Arbitrary shape scene text detection with adaptive text region
  representation.
\newblock In \emph{Proceedings of the IEEE Conference on Computer Vision and
  Pattern Recognition}, pages 6449--6458, 2019{\natexlab{b}}.

\bibitem[Weinman et~al.(2007)Weinman, Learned-Miller, and
  Hanson]{weinman2007fast}
J.~Weinman, E.~Learned-Miller, and A.~Hanson.
\newblock Fast lexicon-based scene text recognition with sparse belief
  propagation.
\newblock In \emph{icdar}, pages 979--983. IEEE, 2007.

\bibitem[Wolf and Jolion(2006)]{wolf2006object}
C.~Wolf and J.-M. Jolion.
\newblock Object count/area graphs for the evaluation of object detection and
  segmentation algorithms.
\newblock \emph{International Journal of Document Analysis and Recognition
  (IJDAR)}, 8\penalty0 (4):\penalty0 280--296, 2006.

\bibitem[Wu et~al.(2019)Wu, Zhang, Liu, Han, Liu, Ding, and Bai]{wu2019editing}
L.~Wu, C.~Zhang, J.~Liu, J.~Han, J.~Liu, E.~Ding, and X.~Bai.
\newblock Editing text in the wild.
\newblock In \emph{Proceedings of the 27th ACM International Conference on
  Multimedia}, pages 1500--1508, 2019.

\bibitem[Wu and Natarajan(2017)]{wu2017self}
Y.~Wu and P.~Natarajan.
\newblock Self-organized text detection with minimal post-processing via border
  learning.
\newblock In \emph{Proceedings of the IEEE Conference on CVPR}, pages
  5000--5009, 2017.

\bibitem[Xia et~al.(2017)Xia, Tian, Wu, Lin, Qin, Yu, and
  Liu]{xia2017deliberation}
Y.~Xia, F.~Tian, L.~Wu, J.~Lin, T.~Qin, N.~Yu, and T.-Y. Liu.
\newblock Deliberation networks: Sequence generation beyond one-pass decoding.
\newblock In \emph{Advances in Neural Information Processing Systems}, pages
  1784--1794, 2017.

\bibitem[Xing et~al.(2019)Xing, Tian, Huang, and Scott]{xing2019convolutional}
L.~Xing, Z.~Tian, W.~Huang, and M.~R. Scott.
\newblock Convolutional character networks.
\newblock In \emph{Proceedings of the IEEE International Conference on Computer
  Vision}, pages 9126--9136, 2019.

\bibitem[Xu et~al.(2015)Xu, Ba, Kiros, Cho, Courville, Salakhudinov, Zemel, and
  Bengio]{xu2015show}
K.~Xu, J.~Ba, R.~Kiros, K.~Cho, A.~Courville, R.~Salakhudinov, R.~Zemel, and
  Y.~Bengio.
\newblock Show, attend and tell: Neural image caption generation with visual
  attention.
\newblock In \emph{International Conference on Machine Learning}, pages
  2048--2057, 2015.

\bibitem[Xue et~al.(2018)Xue, Lu, and Zhan]{xue2018accurate}
C.~Xue, S.~Lu, and F.~Zhan.
\newblock Accurate scene text detection through border semantics awareness and
  bootstrapping.
\newblock In \emph{In Proceedings of European Conference on Computer Vision
  (ECCV)}, 2018.

\bibitem[Yang et~al.(2019)Yang, Guan, Liao, He, Bian, Bai, Yao, and
  Bai]{yang2019symmetry}
M.~Yang, Y.~Guan, M.~Liao, X.~He, K.~Bian, S.~Bai, C.~Yao, and X.~Bai.
\newblock Symmetry-constrained rectification network for scene text
  recognition.
\newblock In \emph{Proceedings of the IEEE International Conference on Computer
  Vision}, pages 9147--9156, 2019.

\bibitem[Yang et~al.(2020)Yang, Jin, Huang, and Lin]{yang2020swaptext}
Q.~Yang, H.~Jin, J.~Huang, and W.~Lin.
\newblock Swaptext: Image based texts transfer in scenes.
\newblock \emph{arXiv preprint arXiv:2003.08152}, 2020.

\bibitem[Yang et~al.(2017)Yang, He, Zhou, Kifer, and Giles]{yang2017learning}
X.~Yang, D.~He, Z.~Zhou, D.~Kifer, and C.~L. Giles.
\newblock Learning to read irregular text with attention mechanisms.
\newblock In \emph{Proceedings of the Twenty-Sixth International Joint
  Conference on Artificial Intelligence, IJCAI-17}, pages 3280--3286, 2017.

\bibitem[Yao et~al.(2014)Yao, Bai, Shi, and Liu]{yao2014strokelets}
C.~Yao, X.~Bai, B.~Shi, and W.~Liu.
\newblock Strokelets: A learned multi-scale representation for scene text
  recognition.
\newblock In \emph{Proceedings of the IEEE Conference on Computer Vision and
  Pattern Recognition (CVPR)}, pages 4042--4049, 2014.

\bibitem[Yao et~al.(2016)Yao, Bai, Sang, Zhou, Zhou, and Cao]{yao2016scene}
C.~Yao, X.~Bai, N.~Sang, X.~Zhou, S.~Zhou, and Z.~Cao.
\newblock Scene text detection via holistic, multi-channel prediction.
\newblock \emph{arXiv preprint arXiv:1606.09002}, 2016.

\bibitem[Ye and Doermann(2015)]{ye2015text}
Q.~Ye and D.~Doermann.
\newblock Text detection and recognition in imagery: A survey.
\newblock \emph{IEEE transactions on pattern analysis and machine
  intelligence}, 37\penalty0 (7):\penalty0 1480--1500, 2015.

\bibitem[Ye et~al.(2003)Ye, Gao, Wang, and Zeng]{ye2003robust}
Q.~Ye, W.~Gao, W.~Wang, and W.~Zeng.
\newblock A robust text detection algorithm in images and video frames.
\newblock \emph{IEEE ICICS-PCM}, pages 802--806, 2003.

\bibitem[Yi and Tian(2011)]{yi2011text}
C.~Yi and Y.~Tian.
\newblock Text string detection from natural scenes by structure-based
  partition and grouping.
\newblock \emph{IEEE Transactions on Image Processing}, 20\penalty0
  (9):\penalty0 2594--2605, 2011.

\bibitem[Yin et~al.(2017)Yin, Wu, Zhang, and Liu]{yin2017scene}
F.~Yin, Y.-C. Wu, X.-Y. Zhang, and C.-L. Liu.
\newblock Scene text recognition with sliding convolutional character models.
\newblock \emph{arXiv preprint arXiv:1709.01727}, 2017.

\bibitem[Yin et~al.(2014)Yin, Yin, Huang, and Hao]{yin2014robust}
X.-C. Yin, X.~Yin, K.~Huang, and H.-W. Hao.
\newblock Robust text detection in natural scene images.
\newblock \emph{IEEE transactions on pattern analysis and machine
  intelligence}, 36\penalty0 (5), 2014.

\bibitem[Yin et~al.(2016)Yin, Zuo, Tian, and Liu]{yin2016text}
X.-C. Yin, Z.-Y. Zuo, S.~Tian, and C.-L. Liu.
\newblock Text detection, tracking and recognition in video: A comprehensive
  survey.
\newblock \emph{IEEE Transactions on Image Processing}, 25\penalty0 (6), 2016.

\bibitem[Yu et~al.(2020)Yu, Li, Zhang, Han, Liu, and Ding]{yu2020towards}
D.~Yu, X.~Li, C.~Zhang, J.~Han, J.~Liu, and E.~Ding.
\newblock Towards accurate scene text recognition with semantic reasoning
  networks.
\newblock \emph{arXiv preprint arXiv:2003.12294}, 2020.

\bibitem[Yuan et~al.(2018)Yuan, Zhu, Xu, Li, and Hu]{yuan2018chinese}
T.-L. Yuan, Z.~Zhu, K.~Xu, C.-J. Li, and S.-M. Hu.
\newblock Chinese text in the wild.
\newblock \emph{arXiv preprint arXiv:1803.00085}, 2018.

\bibitem[Zhan and Lu(2019)]{zhan2019esir}
F.~Zhan and S.~Lu.
\newblock Esir: End-to-end scene text recognition via iterative image
  rectification.
\newblock In \emph{Proceedings of the IEEE Conference on Computer Vision and
  Pattern Recognition}, 2019.

\bibitem[Zhan et~al.(2018)Zhan, Lu, and Xue]{zhan2018verisimilar}
F.~Zhan, S.~Lu, and C.~Xue.
\newblock Verisimilar image synthesis for accurate detection and recognition of
  texts in scenes.
\newblock 2018.

\bibitem[Zhang et~al.(2019)Zhang, Liang, Huang, En, Han, Ding, and
  Ding]{zhang2019look}
C.~Zhang, B.~Liang, Z.~Huang, M.~En, J.~Han, E.~Ding, and X.~Ding.
\newblock Look more than once: An accurate detector for text of arbitrary
  shapes.
\newblock \emph{Proceedings of the IEEE Conference on Computer Vision and
  Pattern Recognition (CVPR)}, 2019.

\bibitem[Zhang and Chang()]{zhang2003bayesian}
D.~Zhang and S.-F. Chang.
\newblock A bayesian framework for fusing multiple word knowledge models in
  videotext recognition.
\newblock In \emph{Computer Vision and Pattern Recognition, 2003.} IEEE.

\bibitem[Zhang et~al.(2018)Zhang, Liu, Jin, and Luo]{ZhangAAAI2018}
S.~Zhang, Y.~Liu, L.~Jin, and C.~Luo.
\newblock Feature enhancement network: A refined scene text detector.
\newblock In \emph{Proceedings of AAAI, 2018}, 2018.

\bibitem[Zhang et~al.(2020)Zhang, Zhu, Hou, Liu, Yang, Wang, and
  Yin]{zhang2020deep}
S.-X. Zhang, X.~Zhu, J.-B. Hou, C.~Liu, C.~Yang, H.~Wang, and X.-C. Yin.
\newblock Deep relational reasoning graph network for arbitrary shape text
  detection.
\newblock \emph{arXiv preprint arXiv:2003.07493}, 2020.

\bibitem[Zhang et~al.(2016)Zhang, Zhang, Shen, Yao, Liu, and
  Bai]{zhang2016multi}
Z.~Zhang, C.~Zhang, W.~Shen, C.~Yao, W.~Liu, and X.~Bai.
\newblock Multi-oriented text detection with fully convolutional networks.
\newblock In \emph{Proceedings of the IEEE Conference on Computer Vision and
  Pattern Recognition (CVPR)}, 2016.

\bibitem[Zhiwei et~al.(2010)Zhiwei, Linlin, and Lim]{zhiwei2010edge}
Z.~Zhiwei, L.~Linlin, and T.~C. Lim.
\newblock Edge based binarization for video text images.
\newblock In \emph{2010 20th International Conference on Pattern Recognition
  (ICPR)}, pages 133--136. IEEE, 2010.

\bibitem[Zhou et~al.(2017)Zhou, Yao, Wen, Wang, Zhou, He, and
  Liang]{Zhou_2017_CVPR}
X.~Zhou, C.~Yao, H.~Wen, Y.~Wang, S.~Zhou, W.~He, and J.~Liang.
\newblock {EAST}: An efficient and accurate scene text detector.
\newblock In \emph{The IEEE Conference on Computer Vision and Pattern
  Recognition (CVPR)}, 2017.

\bibitem[Zhu et~al.(2016)Zhu, Yao, and Bai]{zhu2016scene}
Y.~Zhu, C.~Yao, and X.~Bai.
\newblock Scene text detection and recognition: Recent advances and future
  trends.
\newblock \emph{Frontiers of Computer Science}, 10\penalty0 (1):\penalty0
  19--36, 2016.

\bibitem[Zitnick and Doll{\'a}r(2014)]{zitnick2014edge}
C.~L. Zitnick and P.~Doll{\'a}r.
\newblock Edge boxes: Locating object proposals from edges.
\newblock In \emph{In Proceedings of European Conference on Computer Vision
  (ECCV)}, pages 391--405. Springer, 2014.

\end{thebibliography}


\end{document}